\documentclass{article}

\usepackage[nonatbib,final]{neurips_2024}
\usepackage[square,numbers]{natbib}
\usepackage[utf8]{inputenc} %
\usepackage[T1]{fontenc}    %
\usepackage{hyperref}       %
\usepackage{url}            %
\usepackage{booktabs}       %
\usepackage{amsfonts}       %
\usepackage{nicefrac}       %
\usepackage{microtype}      %
\usepackage{xcolor}         %
\usepackage{xspace}
\usepackage{amsmath}
\usepackage{amssymb}
\usepackage{graphicx}
\usepackage{subcaption}
\usepackage{enumitem}
\usepackage{colortbl}
\usepackage{tablefootnote}
\usepackage{multirow}

\definecolor{darkred}{rgb}{0.9,0.2,0.2}
\definecolor{colordataleak}{gray}{0.9}
\definecolor{colorbest}{rgb}{0.9,0.0,0.0}
\definecolor{colorsecondbest}{rgb}{0.0,0.0,0.9}
\newcommand{\dataleak}{{\cellcolor{colordataleak}}}
\newcommand{\best}[1]{{\color{colorbest}#1}}
\newcommand{\secondbest}[1]{{\color{colorsecondbest}#1}}

\newcommand{\ourmodel}{ReF-LDM\xspace}
\newcommand{\cachekv}{CacheKV\xspace}
\newcommand{\E}{\mathcal{E}}
\newcommand{\D}{\mathcal{D}}

\newcommand{\timeidloss}{timestep-scaled identity loss\xspace}
\newcommand{\Timeidloss}{Timestep-scaled identity loss\xspace}
\newcommand{\ffhqref}{FFHQ-Ref\xspace}
\newcommand{\ffhqrefsevere}{FFHQ-Ref-Severe\xspace}
\newcommand{\ffhqrefmoderate}{FFHQ-Ref-Moderate\xspace}

\newcommand{\loss}{\mathcal{L}}
\newcommand{\lldm}{\loss_\mathrm{LDM}}
\newcommand{\lid}{\loss_\mathrm{ID}}
\newcommand{\ltimeid}{\loss_\mathrm{time\,ID}}
\newcommand{\lambdaid}{\lambda_\mathrm{time\,ID}}
\newcommand{\at}{\bar{\alpha_t}}
\newcommand{\sqrtat}{\sqrt{\at}}
\newcommand{\x}{\textbf{x}}
\newcommand{\xT}{\x_T}

\newcommand{\xlq}{\x_{\text{LQ}}}
\newcommand{\xhq}{\x_{\text{HQ}}}
\newcommand{\xref}{\x_{\text{ref}}}
\newcommand{\xrefs}{\{\xref\}}

\newcommand{\z}{\textbf{z}}
\newcommand{\zT}{\z_T}
\newcommand{\zt}{\z_t}
\newcommand{\zlq}{\z_{\text{LQ}}}
\newcommand{\zref}{\z_{\text{ref}}}
\newcommand{\zrefs}{\{\zref\}}

\title{\ourmodel: A Latent Diffusion Model for Reference-based Face Image Restoration}

\author{
    \textbf{Chi-Wei Hsiao\textsuperscript{1}} \hspace{1em} 
    \textbf{Yu-Lun Liu\textsuperscript{2}} \hspace{1em}
    \textbf{Cheng-Kun Yang\textsuperscript{1}} \hspace{1em} 
    \textbf{Sheng-Po Kuo\textsuperscript{1}} \\
    \textbf{Yucheun Kevin Jou\textsuperscript{1}} \hspace{1em}
    \textbf{Chia-Ping Chen\textsuperscript{1}} \\
    \\
    \textsuperscript{1}MediaTek
    \hspace{1em}
    \textsuperscript{2}National Yang Ming Chiao Tung University
}

\begin{document}

\maketitle

\begin{abstract}
  While recent works on blind face image restoration have successfully produced impressive high-quality (HQ) images with abundant details from low-quality (LQ) input images, the generated content may not accurately reflect the real appearance of a person. To address this problem, incorporating well-shot personal images as additional reference inputs could be a promising strategy. Inspired by the recent success of the Latent Diffusion Model (LDM), we propose \ourmodel---an adaptation of LDM designed to generate HQ face images conditioned on one LQ image and multiple HQ reference images. Our model integrates an effective and efficient mechanism, \cachekv, to leverage the reference images during the generation process. Additionally, we design a \timeidloss, enabling our LDM-based model to focus on learning the discriminating features of human faces. Lastly, we construct \ffhqref, a dataset consisting of 20,405 high-quality (HQ) face images with corresponding reference images, which can serve as both training and evaluation data for reference-based face restoration models.
\end{abstract}

\section{Introduction}
\label{sec:intro}
Recent works~\cite{gu2022vqfr,tsai2024dual,zhou2022towards} have achieved impressive results in generating a realistic high-quality (HQ) face image from an input low-quality (LQ) image. 
However, the important features of a person's face may be corrupted in the LQ image, and thus the reconstructed image may look like a different person.
To tackle this problem, besides the LQ image, additional HQ images of this person can be used as reference input. Moreover, allowing multiple reference images may lead to better quality because they offer more comprehensive appearance of this person in different conditions, e.g., different poses, expressions, or lighting.

A previous work~\cite{li2022learning} has explored using multiple reference images for face restoration. Their method, however, depends on a face landmark model to detect facial components (i.e., eyes, nose, and mouse), which may become unreliable when the input LQ image is severely degraded. 
Besides, latent diffusion model (LDM)~\cite{rombach2022high} has also been used in different image generating tasks with different input conditions, such as low-resolution images, semantic maps, or sketch images~\cite{rombach2022high, zhang2023adding}.

Inspired by the recent success of LDM, we propose \textbf{\ourmodel} for reference-based face image restoration.
Unlike previous conditional LDM methods where their input conditions are usually spatially aligned with the target image,  the reference images are not aligned with the target HQ image in our case.
Therefore, we design a \textbf{\cachekv} mechanism, which effectively and efficiently integrates the reference images, albeit with different poses and expressions. 
Furthermore, we introduce a \timeidloss to drive the reconstructed image to look like the same person of the LQ and reference images.
Lastly, we also construct a new large-scale dataset of face images with corresponding reference images, which can serve as both training and evaluation datasets for future reference-based face restoration research.

With the above components, our \ourmodel outperforms recent state-of-the-art methods with a significant improvement in face identity similarity.
Extensive ablation studies for the proposed \cachekv mechanism and \timeidloss are also conducted and reported.
The main contributions of this work can be summarized as:
\begin{itemize}[noitemsep, topsep=0pt, parsep=4pt, partopsep=0pt, leftmargin=20pt]
  \item We propose \ourmodel, which features an effective and efficient \cachekv mechanism, for restoring an LQ face image using multiple reference images.
  \item We introduce a \timeidloss, which considers the characteristics of diffusion models and helps \ourmodel better learn the discriminating features of human identities.
  \item We construct \ffhqref, a dataset comprising 20,406 high-quality face images and their corresponding reference images, to facilitate the advance of reference-based face image restoration.
\end{itemize}

\begin{figure}[t]
\centering
\scalebox{1.0}{%
\begin{subfigure}{0.33\textwidth}
  \centering
  \includegraphics[width=\linewidth]{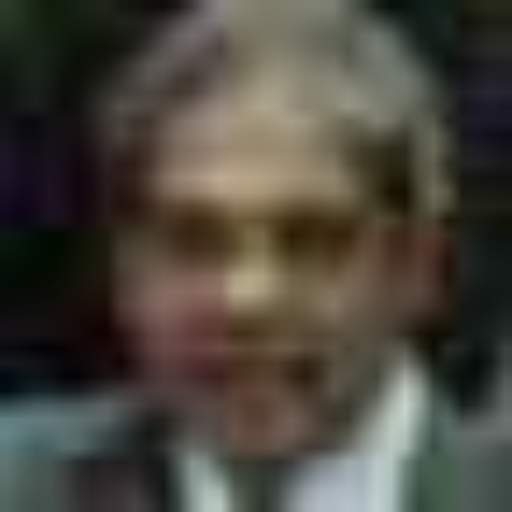}
  \captionsetup{skip=4pt}
  \caption{Input LQ image}
  \label{fig:lq}
\end{subfigure}%
\begin{subfigure}{0.33\textwidth}
  \centering
  \includegraphics[width=\linewidth]{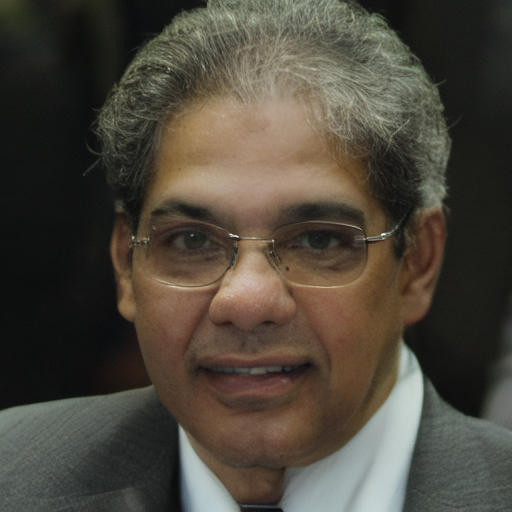}
  \captionsetup{skip=4pt}
  \caption{LDM}
  \label{fig:ldm}
\end{subfigure}%
\begin{subfigure}{0.33\textwidth}
  \centering
  \includegraphics[width=\linewidth]{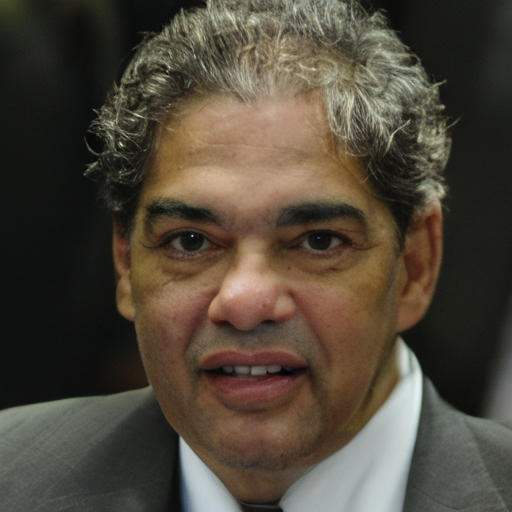}
  \captionsetup{skip=4pt}
  \caption{ReF-LDM}
  \label{fig:ref-ldm}
\end{subfigure}
}
\scalebox{1.0}{%
\begin{subfigure}{.99\textwidth}
  \centering
  \vspace{4pt}
  \includegraphics[width=\linewidth]{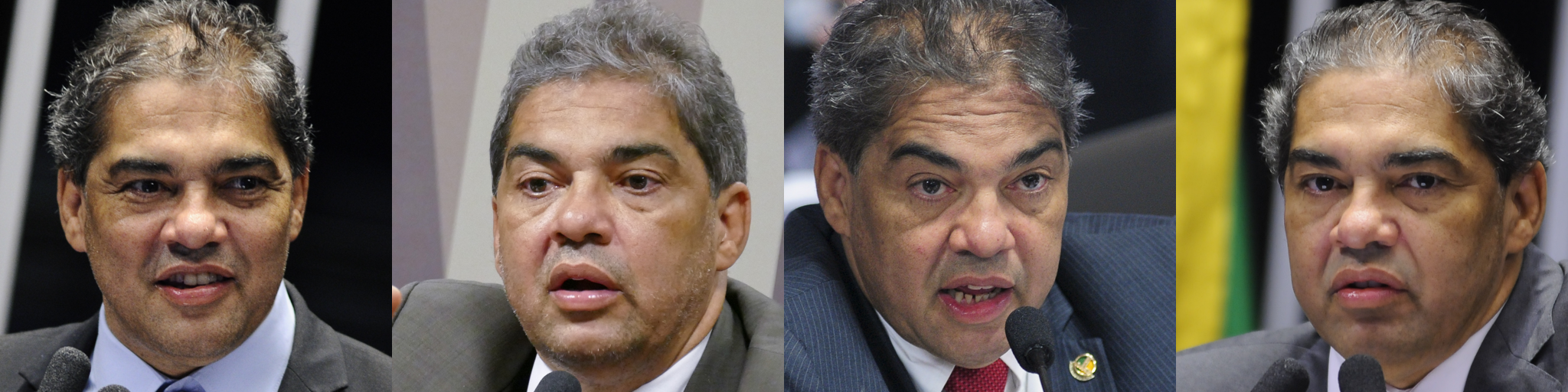}
  \captionsetup{skip=4pt}
  \caption{Input reference images}
  \label{fig:references}
\end{subfigure}
}
\captionsetup{skip=5pt}
\caption{
\textbf{Reference-based face image restoration.}
Given an input low-quality face image (a), a Latent Diffusion Model (LDM) can reconstruct a high-quality image (b); however, it may not be faithful to the individual's facial identity. To address this problem, we propose \ourmodel, which restores a high-quality image with faithful details (c) by utilizing additional reference images (d).
}
\label{fig:teaser}
\end{figure}

\section{Related work}
\paragraph{Face restoration without personal reference images}
Numerous studies have been proposed for blind face image restoration~\cite{wang2021towards,chen2021progressive,gu2022vqfr,zhou2022towards,pouyanfar2023frr,lau2024ented,tsai2024dual}.
Recent works such as VQFR~\cite{gu2022vqfr} and CodeFormer~\cite{zhou2022towards} have achieved promising results by exploiting VQGAN, while DAEFR~\cite{tsai2024dual} further employs a dual-branch encoder to mitigate the domain gap between LQ and HQ images.
Inspired by the success of diffusion models, several works~\cite{saharia2022image,zhao2023towards,lin2023diffbir,wang2024exploiting,suin2024diffuse} have adopted diffusion models for face image restoration.
However, as these methods do not leverage reference images, the restored images may differ from the authentic facial appearance of a person, especially when an input image is severely degraded.

\paragraph{Face restoration with personal reference images}
Several methods~\cite{li2018learning, li2020enhanced, li2022learning,  nitzan2022mystyle} have attempted to utilize additional reference images to enhance personal fidelity in face restoration.
GFRNet~\cite{li2018learning} warps a single reference image to match the face pose of the LQ image, while ASFNet~\cite{li2020enhanced} selects the reference image with the closest matching facial landmarks to serve as the network input.
Closer to the setting of this work, DMDNet~\cite{li2022learning} also utilizes multiple reference images. It detects facial landmarks on the LQ image and the reference images to extract features of facial components, and then integrates these features into the model by querying the corresponding components. However, their method relies on landmark detection, which may not be robust on severely degraded LQ images. In contrast, our \ourmodel implicitly learns the correspondences between the features of the LQ image and the reference images, without the need for landmark detection.
From a different perspective, MyStyle~\cite{nitzan2022mystyle} adopts a per-person optimization setting, leveraging hundreds of images of an individual to define a personalized subspace within the latent space of a StyleGAN~\cite{karras2019style}. In comparison, our approach offers greater flexibility, capable of utilizing one to several reference images without the need for personalized model optimization for each individual.

\paragraph{Latent diffusion models with image conditions}
Previous work demonstrates that LDM can generate an image from a low-resolution image by simple channel-axis concatenation~\cite{rombach2022high}. However, reference images in our task are not spatially aligned with the target HQ image, thus requiring a more sophisticated integration mechanism.
MasaCtrl~\cite{cao_2023_masactrl} achieves text-to-image synthesis with a single reference image by replacing the original keys and values tokens with those from the reference image. However, their solution requires passing the reference image through the denoising network for multiple timesteps, which increases computation and limits its feasibility for extending to multiple reference images.
In contrast, we propose an efficient \cachekv mechanism that leverages multiple reference images by eliminating the redundant network passes.

\section{The proposed \ourmodel model}
\label{sec:method}

In this section, we present the proposed \ourmodel model. We introduce the network architecture in Sec.~\ref{subsec:method_model}, where a \cachekv mechanism is designed to leverage reference images. We illustrate how to train our model with the \timeidloss in Sec.~\ref{subsec:method_loss}.

\begin{figure}[!h]
  \centering
  \includegraphics[width=\linewidth]{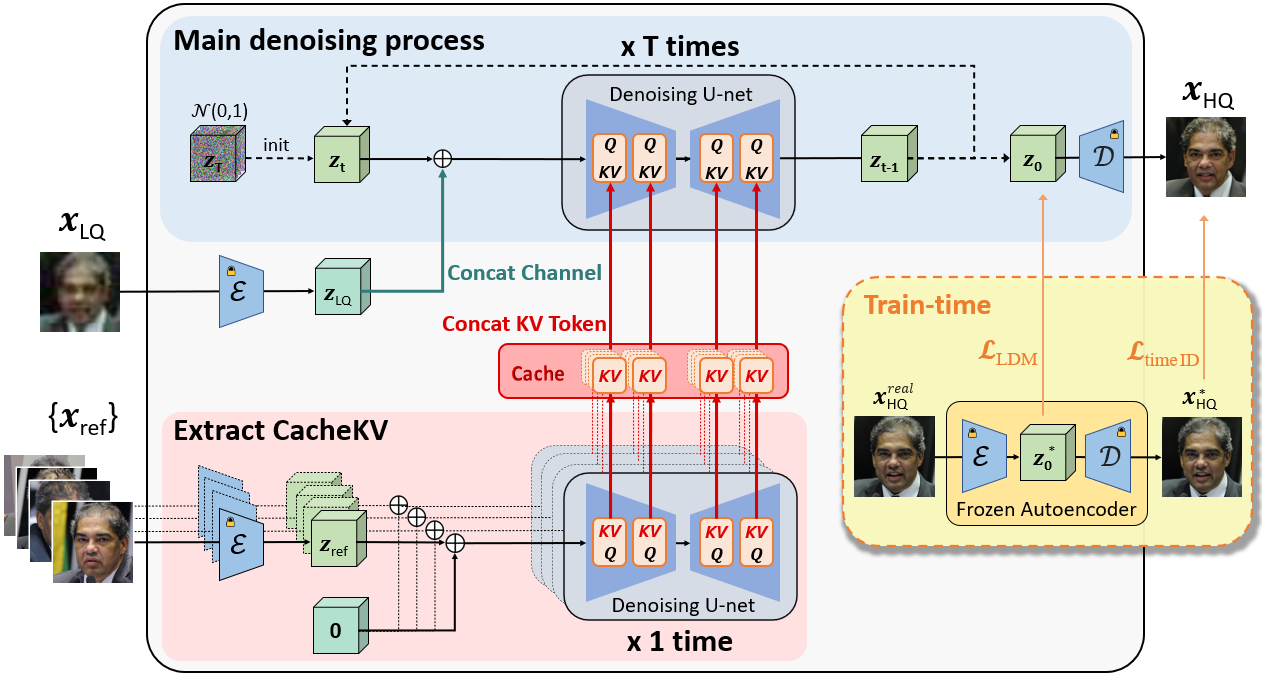}
  \caption{
  \textbf{The proposed \ourmodel pipeline.} Our model accepts a low-quality image and multiple high-quality reference images as input and generates a high-quality image.
  The blue top panel alone represents a typical LDM~\cite{rombach2022high} denoising process.
  For an LQ image $\xlq$, we concatenate its latent $\zlq$ with $\zt$ along the channel axis to serve as the input for the denoising U-net.
  For the reference images $\xrefs$, we design a \textcolor{black}{\textbf{\cachekv}} mechanism, depicted in the red panel, to extract and cache their key and value tokens using the same denoising U-net for just one time. These cached KV tokens can then be utlized repeatedly in each of the $T$ timesteps of the main denoising process.
  During training, we adopt the classic LDM loss ($\lldm$) and introduce a \timeidloss ($\ltimeid$).
  }
  \label{fig:model}
\end{figure}

\subsection{Model architecture of \ourmodel}
\label{subsec:method_model}

The proposed \ourmodel accepts an input LQ image and multiple reference images to generate a target HQ image. Its model architecture is based on the latent diffusion model~\cite{rombach2022high}, with additional designs to incorporate the input LQ image and the reference images.

\subsubsection{Preliminaries on Latent Diffusion Model}
\label{subsubsec:method_prelim}
To generate an image, an image diffusion model~\cite{ho2020denoising} starts from a noisy image $\xT \in \mathbb{R}^{H \times W \times 3}$, initialized with a Gaussian distribution, and gradually denoises it to a clean image $\x_0$ with a denoising network over $T$ timesteps. A latent diffusion model~\cite{rombach2022high} operates similarly, but the diffusion process takes place in a more compact latent space of a pre-trained and frozen autoencoder (encoder $\E$ and decoder $\D$). That is, it begins with a random latent $\zT \in \mathbb{R}^{H_z \times W_z \times C_z}$ and progressively denoises it to a clean latent $\z_0$. A clean image is then generated by passing the clean latent through the decoder during the inference phase, i.e., $\x_0 = \D(\z_0)$; conversely, a ground truth clean latent is obtained by encoding a clean image with the encoder during the training phase, i.e., $\z_0^* = \E(\x_0^*)$.
A typical choice for the denoising network is a U-net with self-attention layers at multiple scales.

\subsubsection{\cachekv: a mechanism for incorporating reference images}
As illustrated in Fig.~\ref{fig:model}, our \ourmodel leverages an input LQ image $\xlq$ and multiple reference images $\xrefs$ to generate a target HQ image $\xhq$.
For an LQ image, we simply concatenate its latent encoded by the frozen encoder, $\zlq=\E(\xlq)$, with the diffusion denoising latent $\zt$ along the channel axis to serve as the input of the denoising U-net. For reference images, we design a \textbf{\cachekv} mechanism.
Essentially, we extract and cache the features of reference images using the same denoising U-net just once; these cached features can then be used repeatedly at each of the $T$ timesteps in the main denoising process.
Specifically, we pass the encoded latent of each reference image, $\zref = \E(\xref)$, through the U-net to extract their keys and values (KVs) at each self-attention layer and store them in a \cachekv.
Subsequently, during the main diffusion process, within each self-attention layer of the U-net, we concatenate the reference KVs (from the corresponding self-attention layer) with the main KVs along the token axis.
This mechanism enables the U-net to incorporate the additional KVs from the reference images into the main denoising process.
When extracting KVs from the reference images, we use a timestep embedding of $t=0$ and pad $\zref$ with a zero tensor to accommodate the additional channels introduced for the LQ image.

To summarize, for inference, we first run the U-net once to extract \cachekv from the reference images; subsequently, we proceed through the main denoising process for $T$ timesteps, during which the U-net integrates $\zlq$ and reference \cachekv.
For training, in each iteration, we first run the U-net to extract \cachekv, and then we run the U-net again to estimate the target latent from a sampled noisy latent $\zt$, incorporating the conditions $\zlq$ and reference \cachekv.

\subsubsection{Comparing \cachekv with other designs}
\label{subsubsec:method_cachekv_other}

\begin{figure}[!h]
\centering
\begin{subfigure}[b]{0.49\textwidth}
    \centering
    \includegraphics[width=\textwidth]{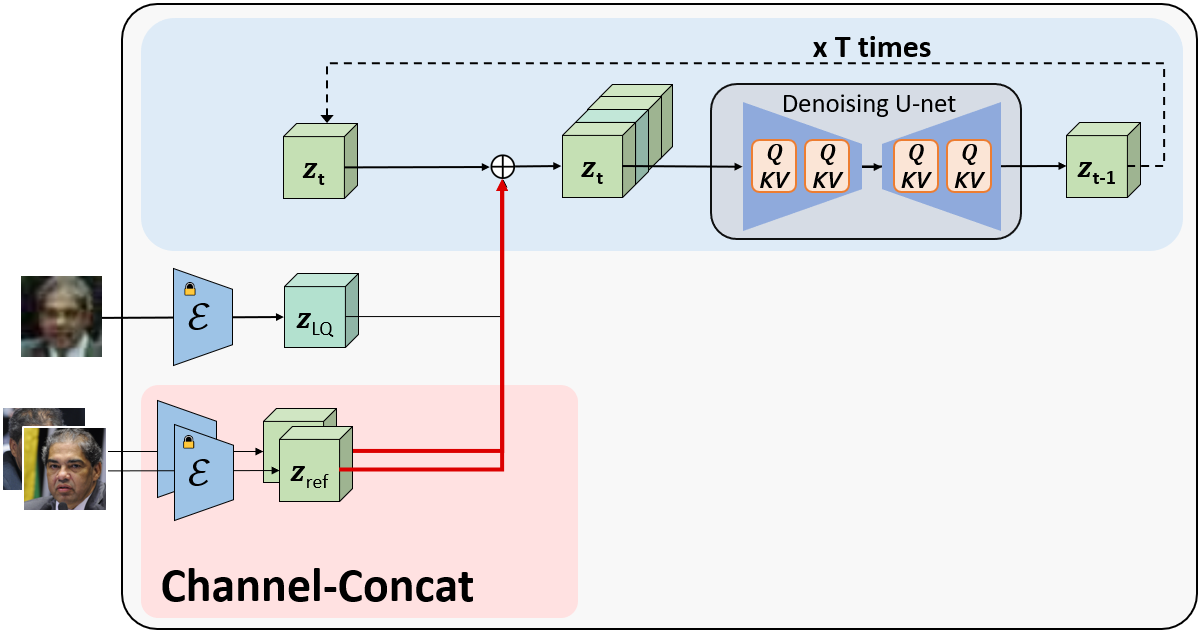}
    \captionsetup{skip=1pt}
    \caption{Channel-concatenation}
    \label{fig:arch_chcat}
\end{subfigure}
\hfill
\begin{subfigure}[b]{0.49\textwidth}
    \centering
    \includegraphics[width=\textwidth]{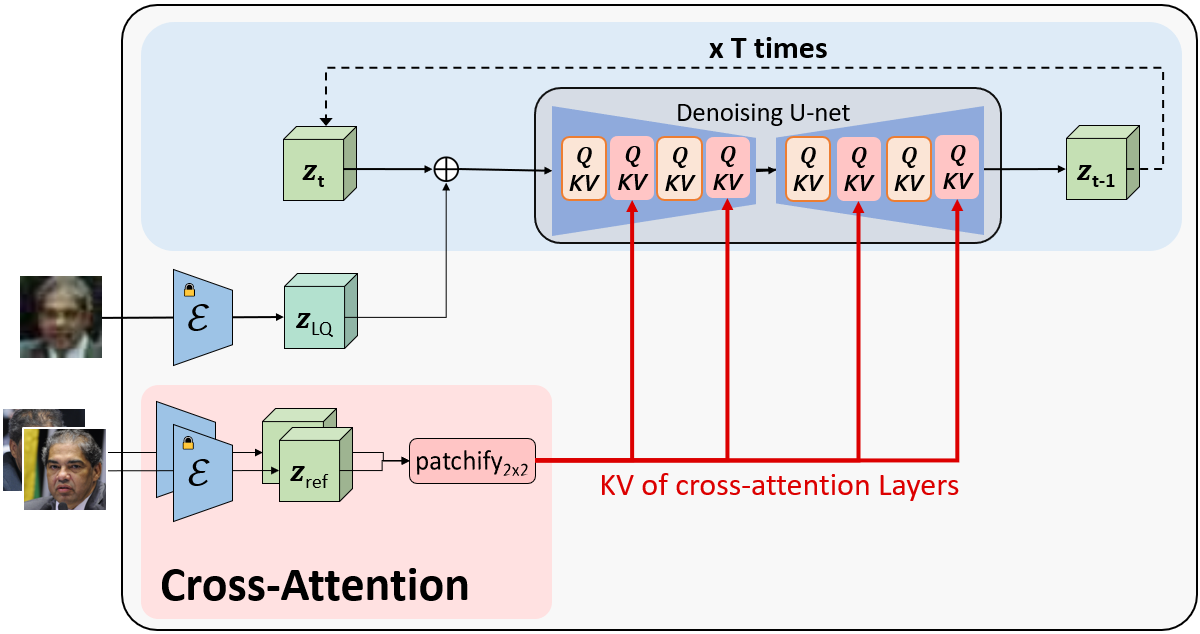}
    \captionsetup{skip=1pt}
    \caption{Cross-attention}
    \label{fig:arch_cross}
\end{subfigure}
\begin{subfigure}[b]{0.49\textwidth}
    \vspace{2pt}
    \centering
    \includegraphics[width=\textwidth]{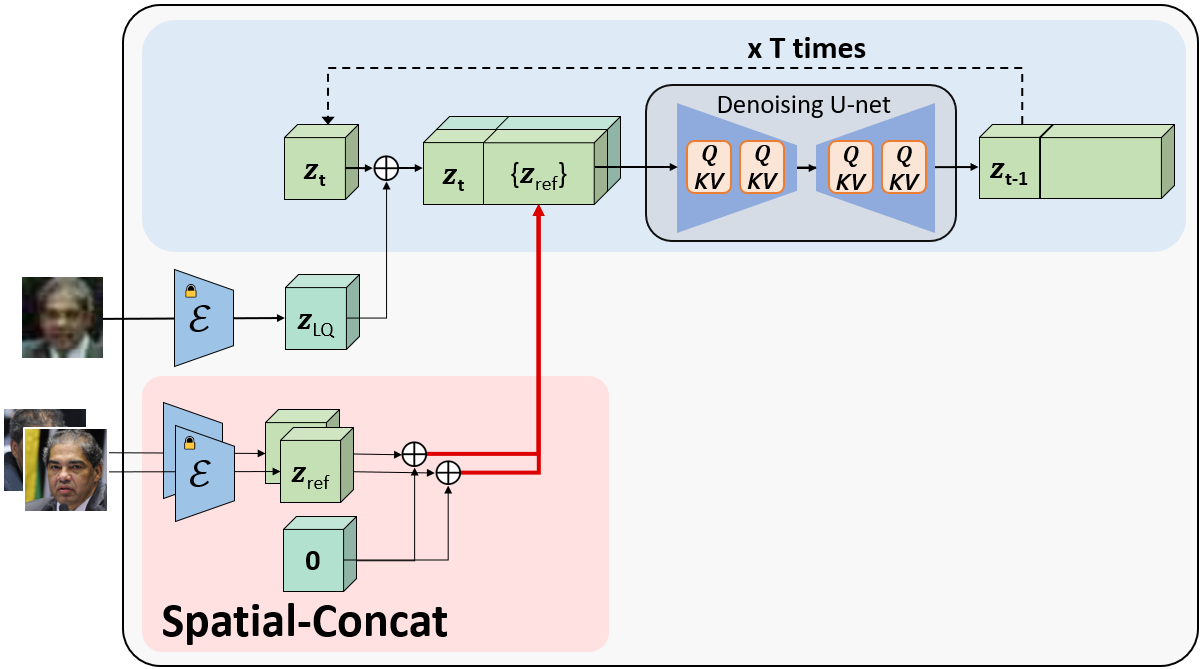}
    \captionsetup{skip=1pt}
    \caption{Spatial-concatenation}
    \label{fig:arch_spcat}
\end{subfigure}
\hfill
\begin{subfigure}[b]{0.49\textwidth}
    \vspace{2pt}
    \centering
    \includegraphics[width=\textwidth]{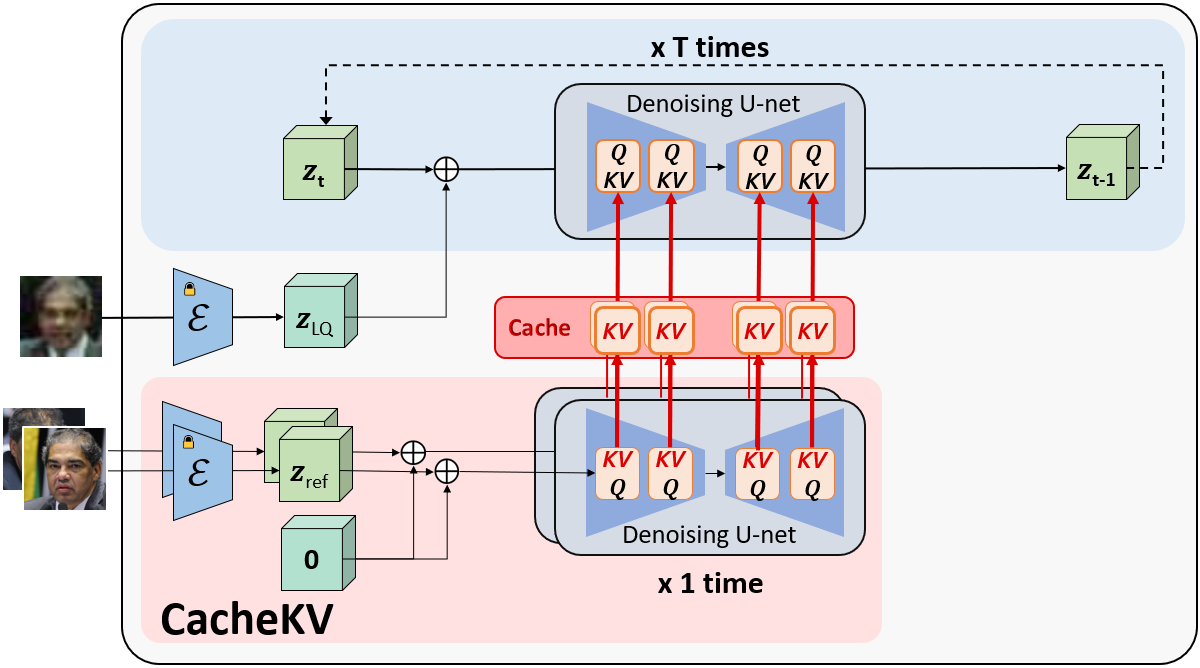}
    \captionsetup{skip=1pt}
    \caption{CacheKV (ours)}
    \label{fig:arch_cachekv}
\end{subfigure}
\caption{Different mechanisms for incorporating reference images into the main denoising process.}
\label{fig:model_other_cachekv}
\end{figure}

There are other intuitive designs for integrating the reference latents $\zrefs$ into the diffusion denoising process. However, they are either ineffective or computationally inefficient compared to the proposed \cachekv.
The quantitative evaluation and computational analysis is reported in Sec.~\ref{subsubsec:abla_cachekv}.
We depict these designs in Fig.~\ref{fig:model_other_cachekv} and provide an intuitive explanation as follows:

\begin{itemize}[noitemsep, topsep=0pt, parsep=4pt, partopsep=0pt, leftmargin=20pt]
\item \textbf{Channel-concatenation}: Concatenating the condition with $\zt$ along the channel axis works well for LQ images (and for other 2D conditions such as semantic maps~\cite{rombach2022high}); however, it is not effective for reference images. A critical difference between these conditions is that—while the LQ image is spatially aligned with the target HQ image, the reference images are not. Therefore, it is challenging for the model to leverage reference images using simple channel-concatenation.
\item \textbf{Cross-attention}: Cross-attention layers have been proven useful for text conditions in text-to-image models~\cite{rombach2022high}. In our ablation experiment, we insert a cross-attention layer after each self-attention layer and use the reference latents $\zrefs$ to produce keys and values. While cross-attention appears to have the potential to address the spatial misalignment problem, it still fails to effectively utilize the reference images. The difference between our \cachekv and the cross-attention setting is that \cachekv provides the reference images in a more aligned feature space for the main denoising process to leverage. Specifically, the \cachekv is extracted using the same U-net and the corresponding self-attention layer as in the main denoising process. In contrast, the cross-attention setting processes the reference images only with the frozen encoder, resulting in features that are less aligned with those in the U-net of the denoising process.
\item \textbf{Spatial-concatenation}: Concatenating $\zrefs$ with $\zt$ along the spatial dimension to serve as the input for U-net also effectively leverages the reference images. Conceptually, spatial-concatenation treats reference images in a very similar way to our \cachekv. In both mechanisms, $\zrefs$ are processed through the denoising U-net, allowing the reference KVs to be accessed by the queries (Qs) of the main diffusion latent $\zt$. However, spatial-concatenation requires significantly more computational resources compared to our \cachekv. It passes $\zrefs$ with $\zt$ to the U-net at each of the $T$ denoising timesteps, whereas \cachekv only passes $\zrefs$ through the U-net once. Moreover, spatial-concatenation also requires significantly more GPU memory, as the spatial size of the input for the U-net increases with the number of reference images. As for a self-attention layer in the U-net, both mechanisms increase memory usage; \cachekv introduces additional reference KVs, while spatial-concatenation introduces reference QKVs.
\end{itemize}

\subsection{\Timeidloss}
\label{subsec:method_loss}

\subsubsection{\Timeidloss}
As this work aims for face image restoration, we employ the identity loss to enhance face similarity, which is adopted in many face-related tasks~\cite{huang2017beyond, richardson2021encoding, wang2021towards}. The identity loss minimizes the distance within the embedding space of a face recognition model, thereby capturing the discriminating features of human faces more effectively than the plain RGB pixel space. In our experiments, we use the ArcFace model~\cite{deng2018arcface} with cosine distance between the 1D embedding vectors as the identity loss.

However, naively adding identity loss to the training of \ourmodel significantly worsens the image quality. One possible explanation might be that, the one-step model prediction $\x_{0|\text{t}} = \D(\z_0|\zt)$ at a very noisy timestep (e.g., $t=T$) is very different from a natural face image and thus out of the distribution that the ArcFace model is trained on; therefore, the identity loss provides ineffective supervision for diffusion models at large timesteps.

Based on this assumption, we propose a \timeidloss, where a timestep-dependent scaling factor is introduced to scale down the identity loss when a larger timestep is sampled in a training step. Specifically, the \timeidloss is defined as:
\begin{equation}
\ltimeid = \sqrtat \cdot \lid = \sqrtat \cdot \Big( 1 - \frac{R(\x) \cdot R(\x^*)}{\|R(\x)\| \|R(\x^*)\|} \Big),
\end{equation}
where $R$ is a face recognition model, and $\sqrtat$ follows the definition in a typical diffusion process~\cite{ho2020denoising,rombach2022high} in which a noisy latent $\zt$ is sampled given a clean latent $\z_0^*$ as:
\begin{equation}
q(\zt|\z_0^*) = \mathcal{N}( \sqrtat \z_0^*, (1 - \at) \mathrm{I})
\end{equation}

\subsubsection{Training \ourmodel with \timeidloss}
We train our \ourmodel with the classic LDM loss and the proposed \timeidloss:
\begin{equation} \loss_{total} = \lldm + \lambdaid \, \ltimeid \end{equation}
Recall that the denoising U-net estimates the target latent in the latent space of the frozen autoencoder, and a typical $\lldm$ is computed as the L1 distance between the estimated latent and the target latent. To compute the identity loss with the face recognition model, which accepts an image as input, we decode the estimated latent into the image space using the frozen decoder,  i.e, $\x_0 = \D(\z_0)$. The experiments in Sec.~\ref{subsubsubsec:abla_timeidloss} show that \timeidloss can improve face similarity without degrading image quality, unlike the naive usage of identity loss.

\section{\ffhqref dataset}
\label{sec:dataset}
Recent works for non-reference-based face restoration commonly train their models with FFHQ dataset~\cite{karras2019style}, which comprises 70,000 high-quality face images of wide appearance variety with appropriate licenses crawled from Flickr. These images are not provided with reference labels originally; however, we find that a good portion of the images are of the same identities. Thus, we construct a reference-based dataset---\ffhqref---based on the FFHQ dataset, with careful consideration described as follows.

\subsection{Finding reference images of the same identity}
\label{subsec:dataset_find}
To determine whether two images belong to the same identity, we utilize the face recognition model ArcFace~\cite{deng2018arcface}. Specifically, we first extract the 1D ArcFace embeddings for all images. Then, for each image, we compute the cosine distances between its embedding and the embeddings of all other images. A distance less than a threshold $r=0.4$ indicates that the images are valid references belonging to the same person. Following this procedure, we identify 20,405 images with corresponding reference images.

\subsection{Splitting data according to identity}
\label{subsec:dataset_split}
To enable the \ffhqref dataset to serve as both training and evaluation datasets for reference-based face restoration models, we divide the images into train, validation, and test splits. However, random data splitting may result in the train and test splits containing images of the same individual, which is not ideal for a fair evaluation. To ensure that all images of a single identity are assigned to only one data split, we group the images based on their identities. Specifically, we consider identity grouping as a graph problem, where each image acts as a vertex and any pair of images with a distance less than $r$ are connected by edges. We then apply the connected component algorithm from graph theory, where each connected component represents a group of images belonging to the same person. Finally, we identified 6,523 identities and divided them into three splits: a train split with 18,816 images of 6,073 identities,  a validation split with 732 images of 300 identities , and a test split with 857 images of 150 identities. We report more statistics in Appendix~\ref{sec:appendix_ffhqref_stats}.

\subsection{Constructing evaluation dataset with practical considerations}
\label{subsec:dataset_test}

\paragraph{Practical Considerations}
For a fair and meaningful evaluation, the input reference images should not be excessively similar to the target image; hence, we set a minimum cosine distance threshold of 0.1 for the test set. Additionally, we manually check the images in the test split to verify that all reference images indeed correspond to the same identity. Furthermore, in the context of reference-based face restoration applications, it is preferable to select input reference images that capture a more comprehensive representation of a person's appearance, such as varying face poses or expressions.
Although a target image in the test split of our \ffhqref may have %
two to nine reference images, different reference-based methods may have their own constraints on the maximum number of input reference images. To emulate a more representative set of reference images, we sort all available reference images of a target image using farthest point sampling on the ArcFace distance.

\paragraph{Degradation synthesis for input LQ images}
For synthesizing input LQ images from ground truth HQ images, we follow the degradation model used in previous works~\cite{wang2021towards, gu2022vqfr,zhou2022towards}:
\begin{equation}
\xlq = \{[( \xhq * k_\sigma )\downarrow_r + n_{\delta}]_{\text{JPEG}_q}\}\uparrow_r,
\end{equation}
where an HQ image is blurred with a Gaussian kernel $k_\sigma$, downsampled by $r$ scale, added with a Gaussian noise $n_{\delta}$, compressed with JPEG quality level $q$, and  upscaled  to the original size.

We construct two evaluation datasets with different degradation levels:
\begin{itemize}[noitemsep, topsep=0pt, parsep=4pt, partopsep=0pt, leftmargin=20pt]
    \item \ffhqrefmoderate: $\sigma$, $r$, $\delta$, and $q$ are sampled from $[0,8]$, $[1,8]$, $[0,15]$, and $[60,100]$.
    \item \ffhqrefsevere: $\sigma$, $r$, $\delta$, and $q$ are sampled from $[8,16]$, $[8,32]$, $[0,20]$, and $[30,100]$.
\end{itemize}

\subsection{Comparison between \ffhqref and existing datasets}
\label{subsec:dataset_compare}
Table~\ref{tab:dataset_comparison} summarizes the differences between our proposed \ffhqref and existing datasets.
While the CelebRef-HQ dataset~\cite{li2022learning} has been constructed to train and evaluate reference-based face restoration models, our \ffhqref dataset contains twice as many images and six times the number of identities compared to CelebRef-HQ. Moreover, built upon FFHQ~\cite{karras2019style}, \ffhqref provides superior image quality over CelebRef-HQ, as indicated by the lower NIQE score (3.68 vs. 3.97). Some ground-truth images in CelebRef-HQ are affected by watermarks and mirror padding artifacts, as shown in Appendix~\ref{sec:appendix_celebrefhq_bad_images}. 

\begin{table}[!h]
\caption{Comparison between the proposed \ffhqref and existing datasets.}
\label{tab:dataset_comparison}
\centering
\begin{tabular}{lccccc}
\toprule
Dataset & With reference & Licensed & Quality & Images & Identities \\
\midrule
FFHQ~\cite{karras2019style} &  & \checkmark & \checkmark & 70,000 & -\\
\midrule
CelebRef-HQ~\cite{li2022learning} & \checkmark & & & 10,555 & 1,005 \\
\textbf{\ffhqref} & \checkmark & \checkmark & \checkmark & \textbf{20,405} & \textbf{6,523} \\
\bottomrule
\end{tabular}
\end{table}

\section{Experiments}

\label{sec:exp}
In this section, we describe the experimental setup in Sec.~\ref{subsec:exp_setup}, discuss ablation studies in Sec.~\ref{subsec:ablation}, and provide the comparison between our \ourmodel and the state-of-the-art methods in Sec.~\ref{subsec:sota}

\subsection{Experimental setup}
\label{subsec:exp_setup}

\subsubsection{Implementation details}
\label{subsubsec:exp_implement}

To exploit more ground truth images without available reference images, we use 68,411 images in the FFHQ dataset to train a VQGAN~\cite{esser2021taming} as the frozen autoencoder and an LDM with only LQ condition. We then finetune our \ourmodel from the LQ-conditioned LDM with the 18,816 images in our \ffhqref dataset. All models are trained excluding the test split images to ensure fair evaluation on our \ffhqref benchmark. In our experiments, we adopt a 512x512 image resolution, fix the number of reference images to five, and set loss scale $\lambdaid$ to 0.1.
During training, we synthesize input LQ images with $\sigma$, $r$, $\delta$, and $q$ sampled from $[0,16]$, $[1,32]$, $[0,20]$, and $[30,100]$, respectively. For inference, we use 100 DDIM~\cite{song2020denoising} steps and a classifier-free-guidance~\cite{ho2022classifier} with a scale of 1.5 towards reference images. We provides more implementation details in the Appendix~\ref{sec:appendix_more_impl}.

\subsubsection{Evaluation datasets and metrics}
\label{subsubsec:exp_dataset_metric}
For evaluation datasets, we use the test split of our \ffhqref with two different degradation levels: severe and moderate. In addition, previous non-reference-based methods commonly use CelebA-Test~\cite{wang2021towards} for evaluation, which comprises 3,000 LQ and HQ image pairs sampled from the CelebA-HQ dataset~\cite{karras2017progressive}. Therefore, we follow the same procedures described in Sec.~\ref{sec:dataset} to construct a subset of 2,533 images with available reference images, termed CelebA-Test-Ref.

For evaluation metrics, we adopt the identity similarity (IDS)~\cite{gu2022vqfr,zhou2022towards}, which is the cosine similarity calculated using the face recognition model ArcFace~\cite{deng2018arcface}. We also use the widely used perceptual metrics LPIPS~\cite{zhang2018perceptual}. As face pixels are more of concern in the task of face restoration, we also measure the face-region LPIPS (fLPIPS), which is the LPIPS calculated using only the pixels in face regions. For assessing no-reference image quality, we adopt NIQE~\cite{mittal2012making}. Furthermore, we measure the FID~\cite{heusel2017gans}, using 70,000 images from the FFHQ dataset as the target distribution.

\subsection{Ablation studies}
\label{subsec:ablation}

We provide the ablation studies of the proposed \cachekv, \timeidloss, and the number of input reference images. In each ablation experiment, we fine-tune the model for 50,000 steps from the same LDM pre-trained without reference images. We compare the difference settings with the \ffhqrefsevere dataset.

\subsubsection{\cachekv and other mechanisms}
\label{subsubsec:abla_cachekv}
The \cachekv is proposed for integrating the input reference images into the diffusion denoising process. We compare it with other mechanisms illustrated in Sec.~\ref{subsubsec:method_cachekv_other}. According to Table~\ref{tab:abla_cachekv}, channel-concatenation and cross-attention fail to leverage reference images to improve the identity similarity (IDS). In contrast, both spatial-concatenation and our \cachekv significantly enhance IDS. Moreover, our \cachekv is more computationally efficient than spatial-concatenation, requiring only 20\% of the inference time and 39\% of the GPU memory.

\begin{table}[!h]
\centering
\caption{Comparison between \cachekv and other mechanisms for input reference images (run with five reference images on a single GTX 1080).}
\label{tab:abla_cachekv}
\begin{tabular}{l|ccc|rr}
\toprule
 & IDS↑ & NIQE↓ & LPIPS↓ & Inference time↓ & Memory↓ \\
\midrule
Channel-concatenation & 0.23 & 4.49 & 0.46 & 4.17 & 1.77 \\
Cross-attention & 0.23 & 4.56 & 0.46 & 14.54 & 2.80 \\
Spatial-concatenation & 0.69& 4.84 & 0.43 & 58.36 & 7.44 \\
\textbf{\cachekv} & 0.65 & 4.38& 0.43 & 12.15 & 2.87 \\
\bottomrule
\end{tabular}
\end{table}

\begin{figure}[!h]
\centering
\begin{minipage}[h!]{0.44\textwidth}
  \centering
  \captionof{table}{Ablation results for the \timeidloss.}
  \label{tab:abla_timeidloss}
  \begin{tabular}{l|cc}
  \toprule
  Loss & IDS↑ & NIQE↓ \\
  \midrule
  $\lldm$ & 0.52 & 4.56 \\
  $\lldm + \lid$ & 0.69  & 6.56 \\
  $\lldm + \ltimeid$ & 0.65 & 4.38 \\
  \bottomrule
  \end{tabular}
  
  \vspace{5pt}
   \captionof{table}{Design choices for ID loss scaling.}
   \label{tab:abla_timeidloss_variants}
  \begin{tabular}{l|cc}
  \toprule
  Scale for ID loss & IDS↑ & NIQE↓ \\
  \midrule
   $\sqrtat$ & 0.65 & 4.38 \\
   $\mathbf{1}_{t < 100}$ & 0.52 & 4.55 \\
   $\mathbf{1}_{t < 500}$ & 0.61 & 4.44 \\
  \bottomrule
  \end{tabular}
\end{minipage}
\hfill
\setcounter{subfigure}{0}
\begin{minipage}[h!]{0.52\textwidth}
  \centering
  \begin{subfigure}{\linewidth}
    \centering
    \includegraphics[width=0.96\linewidth]{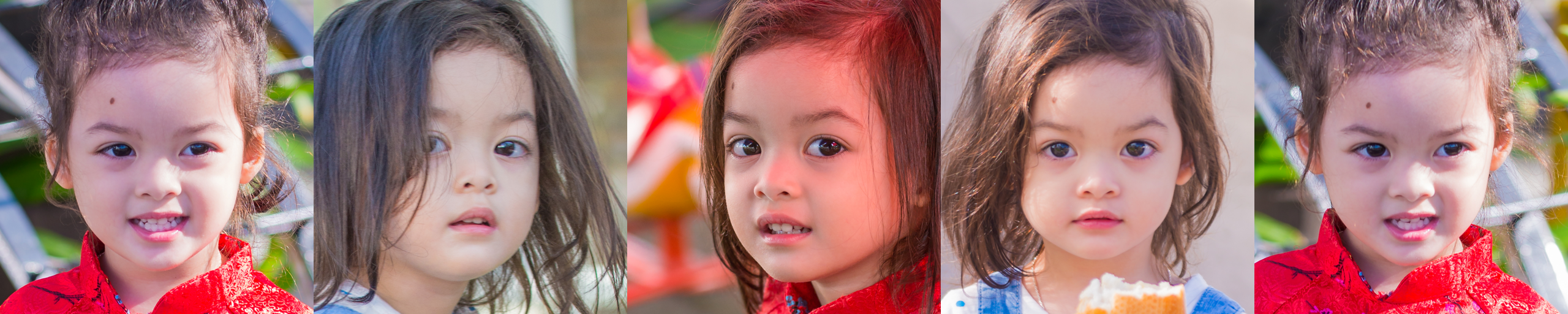}
    \caption{Input references}
  \end{subfigure}
  
  \vspace{0.5em}
  \begin{subfigure}{0.24\textwidth}
    \centering
    \includegraphics[width=.8\linewidth]{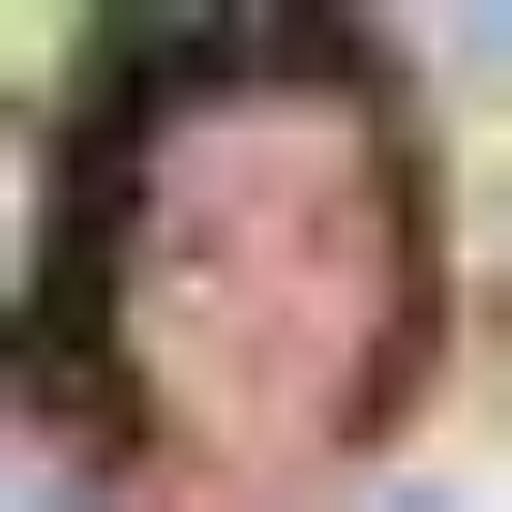}
    \caption{Input LQ}
  \end{subfigure}
  \begin{subfigure}{0.24\textwidth}
    \centering
    \includegraphics[width=.8\linewidth]{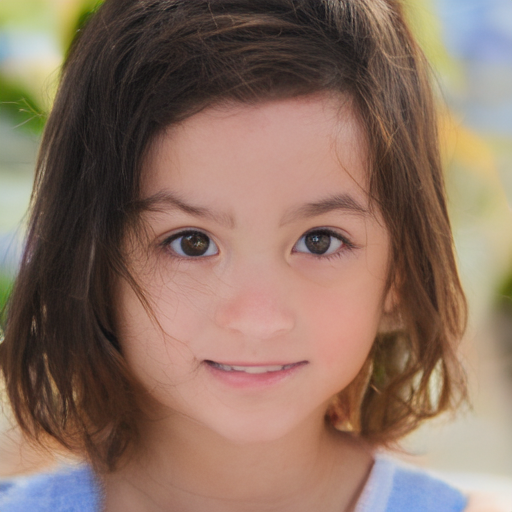}
    \caption{$\lldm$}
  \end{subfigure}
  \begin{subfigure}{0.24\textwidth}
    \centering
    \includegraphics[width=.8\linewidth]{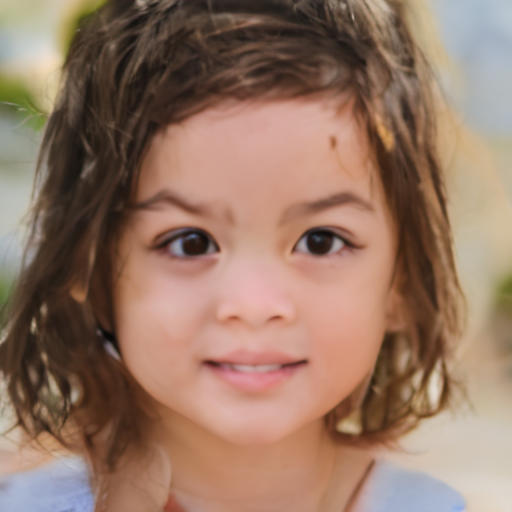}
    \caption{$+\lid$}
  \end{subfigure}
  \begin{subfigure}{0.24\textwidth}
    \centering
    \includegraphics[width=.8\linewidth]{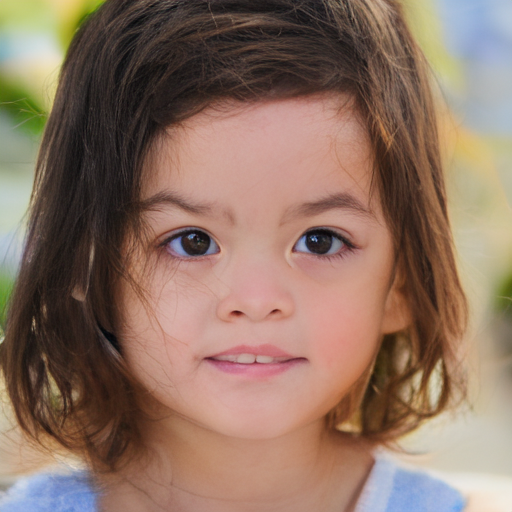}
    \caption{$+\ltimeid$}
  \end{subfigure}
  \vspace{0.5em}
  \caption{Visual ablation results for the \timeidloss.}
  \label{fig:abla_timeidloss}
\end{minipage}
\end{figure}

\subsubsection{\Timeidloss}
\label{subsubsubsec:abla_timeidloss}
To validate the benefits of the proposed \timeidloss, we train \ourmodel with three different loss settings: without identity loss ($\lldm$), with naive identity loss ($\lldm + \lid$), and with the proposed \timeidloss ($\lldm + \ltimeid$). As show in Table~\ref{tab:abla_timeidloss} and Fig.~\ref{fig:abla_timeidloss}, while the naive identity loss can improve identity similarity (IDS), our \timeidloss can do so without sacrificing the image quality (NIQE).

As explained in Sec.~\ref{subsec:method_loss}, we employ $\sqrtat$ to scale down the identity loss for a larger and noisier timestep $t$. In Table~\ref{tab:abla_timeidloss_variants}, we compare this design choice with other alternative scaling factors, $\mathbf{1}_{t < 100}$ and $\mathbf{1}_{\\{t < 500\\}}$, which apply the identity loss only when the sampled timestep $t$ is smaller than $100$ or $500$, respectively. The results suggest that $\sqrtat$ is more effective than the alternatives.

\subsubsection{Multiple input reference images}
\label{subsubsec:abla_nref}
There are two to nine reference images for a target image in the test split of our \ffhqref. While we fix the number of reference images to five when training \ourmodel, the proposed \cachekv mechanism has the flexibility to take varying number of reference images during inference. To validate the effectiveness of utilizing multiple reference images, we evaluate \ourmodel with a maximum of 1, 3, 5, and 8 reference images, respectively. As shown in Table~\ref{tab:abla_nref}, using more reference images significantly improves the identity similarity (from 0.52 to 0.66). However, increasing the number of reference images also increases the computation time, as shown in Table~\ref{tab:abla_nref_time}. Since using eight reference images encounters an out-of-memory issue on a single GTX 1080, we use at most five reference images in our experiments for simplicity.

\begin{table}[!h]
\centering
\begin{minipage}[t]{0.48\textwidth}
\centering
\caption{Image quality with different numbrs of reference images.}
\label{tab:abla_nref}
\begin{tabular}{ccc}
\toprule
Max num refs & IDS↑ & LPIPS↓ \\
\midrule
1 & 0.52 & 0.45 \\
3 & 0.62 & 0.44 \\
5 & 0.65 & 0.43 \\
8 & 0.66 & 0.43 \\
\bottomrule
\end{tabular}
\end{minipage}%
\hfill
\begin{minipage}[t]{0.48\textwidth}
\centering
\caption{Inference time with different numbers of reference images.}
\label{tab:abla_nref_time}
\begin{tabular}{cccc}
\toprule
Num refs & Time@1080↓ & Time@3090↓ \\
\midrule
1 & 4.86 & 3.00 \\
3 & 7.09 & 3.54 \\
5 & 12.03 & 4.51 \\
8 & out-of-memory & 6.26 \\
\bottomrule
\end{tabular}
\end{minipage}
\end{table}

\subsection{Comparison with state-of-the-art methods}
\label{subsec:sota}

\begin{table}[!h]
\centering
\caption{
Comparison of \ourmodel with state-of-the-art methods across three benchmarks.
Note the highlighting \textcolor{colorbest}{1st}, \textcolor{colorsecondbest}{2nd}, and a gray cell indicating \colorbox{colordataleak}{evaluation data leakage for prior methods}.
}
\label{tab:sota_table}
{\resizebox{1\textwidth}{!}{%
\begin{tabular}{@{}l|c@{\hspace{6pt}}c@{\hspace{5pt}}c@{\hspace{5pt}}c@{\hspace{5pt}}|c@{\hspace{6pt}}c@{\hspace{5pt}}c@{\hspace{5pt}}c@{\hspace{5pt}}|c@{\hspace{6pt}}c@{\hspace{5pt}}c@{}}
\toprule
 & \multicolumn{4}{c|}{\ffhqrefsevere} & \multicolumn{4}{c|}{\ffhqrefmoderate} & \multicolumn{3}{c}{CelebA-Test-Ref} \\
\addlinespace
 & \footnotesize IDS↑ & \footnotesize fLPIPS↓ & \footnotesize LPIPS↓ & \footnotesize FID↓ 
 & \footnotesize IDS↑ & \footnotesize fLPIPS↓ & \footnotesize LPIPS↓ & \footnotesize FID↓ 
 & \footnotesize IDS↑ & \footnotesize fLPIPS↓ & \footnotesize LPIPS↓ \\

\midrule
CodeFormer~\cite{zhou2022towards} & \secondbest{\dataleak{0.323}} & \best{\dataleak{0.108}} & \best{\dataleak{0.398}} & \dataleak{51.51} & \dataleak{0.760} & \best{\dataleak{0.084}} & \best{\dataleak{0.301}} & \dataleak{38.78} & 0.660 & \best{0.092} & \best{0.340} \\
VQFR~\cite{gu2022vqfr} & \dataleak{0.308} & \dataleak{0.112} & \secondbest{\dataleak{0.415}} & \dataleak{52.96} & \dataleak{0.659} & \dataleak{0.089} & \secondbest{\dataleak{0.324}} & \dataleak{36.77} & 0.558 & 0.096 & \secondbest{0.352} \\
DAEFR~\cite{tsai2024dual} & \dataleak{0.294} & \dataleak{0.118} & \dataleak{0.435} & \dataleak{49.08} & \dataleak{0.614} & \dataleak{0.093} & \dataleak{0.333} & \dataleak{33.86} & 0.491 & 0.101 & 0.367 \\
LDM & 0.231 & 0.125 & 0.453 & \best{34.40} & 0.753 & 0.095 & 0.344 & \best{32.16} & 0.663 & 0.093 & 0.368 \\
\midrule
DMDNet~\cite{li2022learning}\textsuperscript{\textdagger} & \dataleak{0.185} & \dataleak{0.162} & \dataleak{0.511} & \dataleak{72.66} & \secondbest{\dataleak{0.810}} & \dataleak{0.096} & \dataleak{0.348} & \dataleak{36.60} & \secondbest{0.752} & 0.097 & 0.362 \\
\ourmodel & \best{0.676} & \secondbest{0.110} & 0.429 & \secondbest{37.60} & \best{0.840} & \secondbest{0.088} & 0.332 & \secondbest{33.05} & \best{0.779} & \secondbest{0.093} & 0.368 \\
\bottomrule
\end{tabular}
}}
{\raggedright\scriptsize\textsuperscript{\textdagger}As DMDNet encounters landmark detection failures and fails to yield results for 214/857, 29/857, and 488/2,533 images on the three benchmarks respectively, we compute the metrics for DMDNet using the remaining images.\par}
\end{table}

\subsubsection{Quantitative comparison}
\label{subsubsec:sota_quantitative}

We compare our \ourmodel with state-of-the-art methods on \ffhqrefsevere, \ffhqrefmoderate, and CelebA-Test-Ref.
Table~\ref{tab:sota_table} reports the performance of competing methods in terms of IDS, fLPIPS, LPIPS, and FID (targeting the FFHQ image distribution).
Without the information in reference images, the existing non-reference-based restoration methods (CodeFormer~\cite{zhou2022towards}, VQFR~\cite{gu2022vqfr}, and DAEFR~\cite{tsai2024dual}) fail to preserve the facial identity, leading to significantly lower IDS.
The reference-based method, DMDNet~\cite{li2022learning}, fails to restore the severely degraded images because it depends on unreliable facial landmark detection, reflected by higher fLPIPS.
In contrast, our \ourmodel consistently outperforms DMDNet on identity similarity and other metrics, owing to the proposed \cachekv mechanism and \timeidloss, which effectively leverage the input reference images without the need for landmark detection.
We also note that our method exhibits slightly inferior results in LPIPS metric. 
This is due to the difference in the background pixels, we provide further details in the Appendix~\ref{sec:appendix_background_diff}.
It is also worth mentioning that the competing methods benefit from data leakage on the \ffhqref benchmarks, as their models are trained with the entire FFHQ dataset or with a different train split than the identity-based one in the proposed \ffhqref.

\subsubsection{Qualitative comparison}
\label{subsubsec:sota_qualitative}
\begin{figure}[!h]
\centering
\setlength{\tabcolsep}{1pt} %
\renewcommand{\arraystretch}{0.6} %
\def\mywidth{0.16\linewidth}
\begin{tabular}{cccccc}
\includegraphics[width=\mywidth]{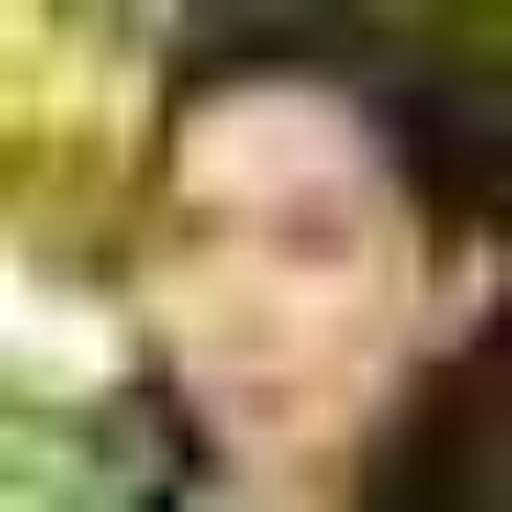} &
\includegraphics[width=\mywidth]{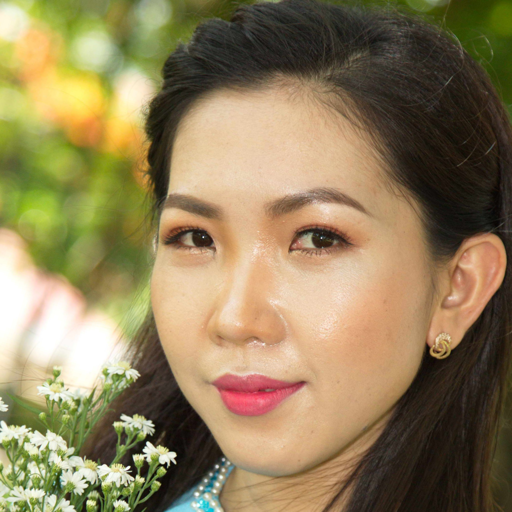} &
\includegraphics[width=\mywidth]{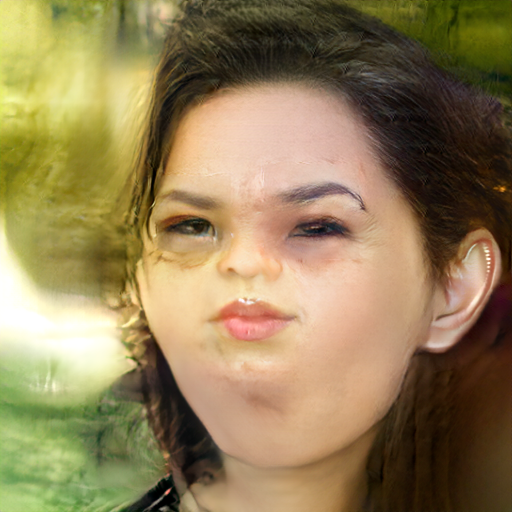} &
\includegraphics[width=\mywidth]{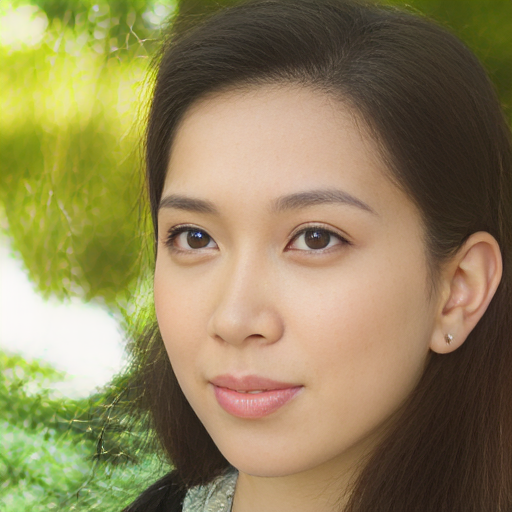} &
\includegraphics[width=\mywidth]{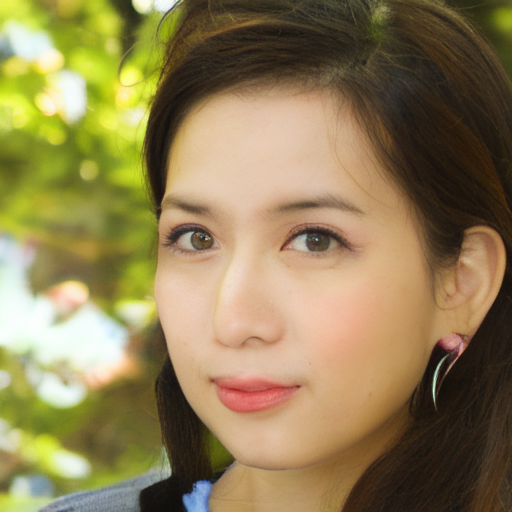} &
\includegraphics[width=\mywidth]{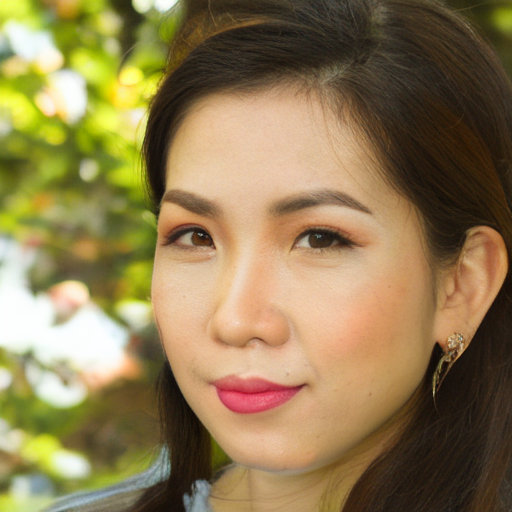} \\
\includegraphics[width=\mywidth]{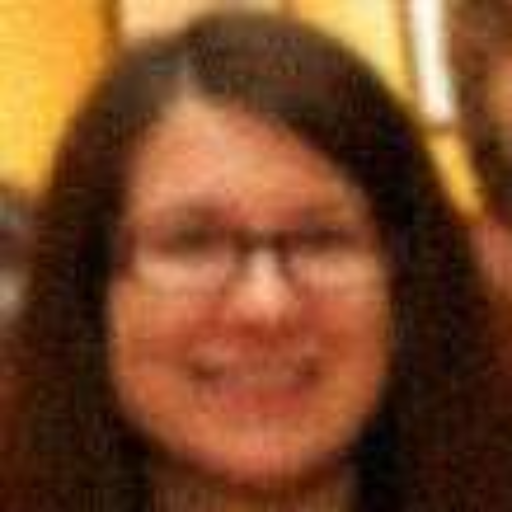} &
\includegraphics[width=\mywidth]{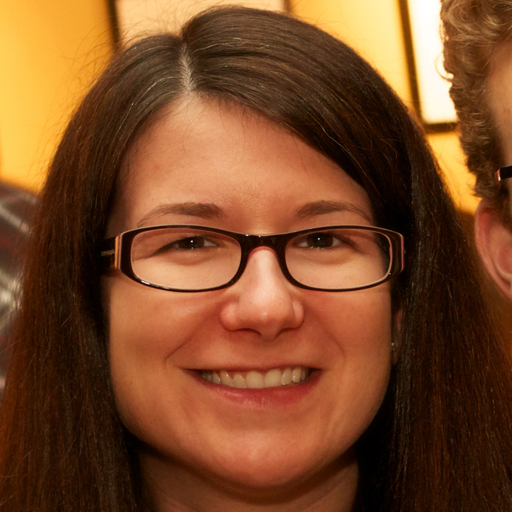} &
\includegraphics[width=\mywidth]{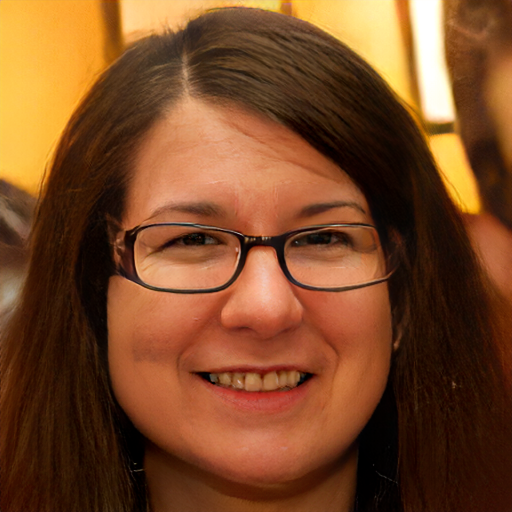} &
\includegraphics[width=\mywidth]{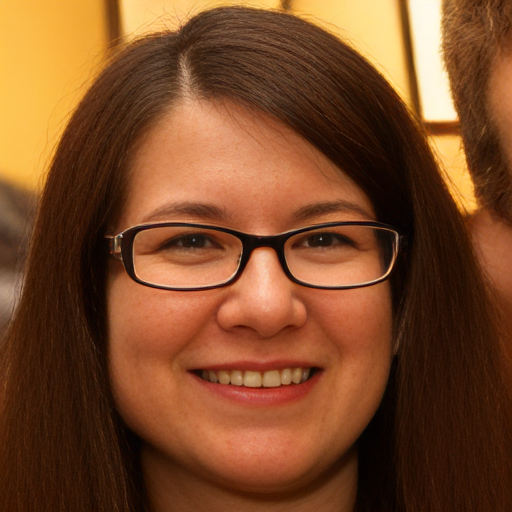} &
\includegraphics[width=\mywidth]{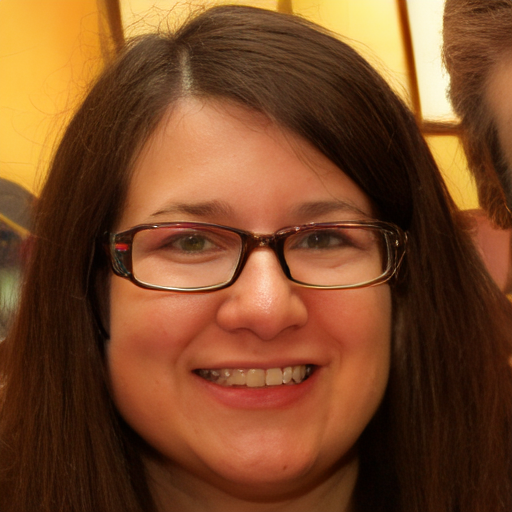} &
\includegraphics[width=\mywidth]{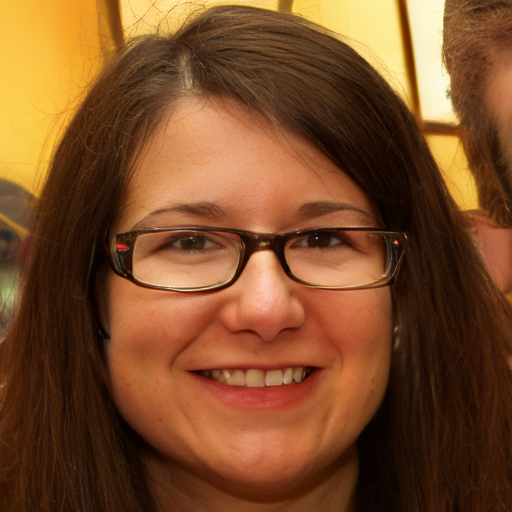} \\
\includegraphics[width=\mywidth]{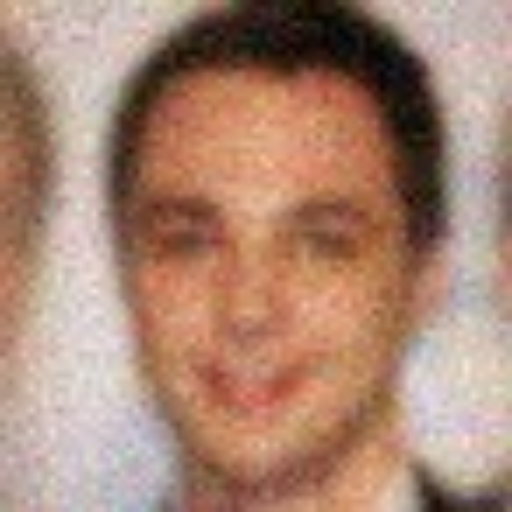} &
\includegraphics[width=\mywidth]{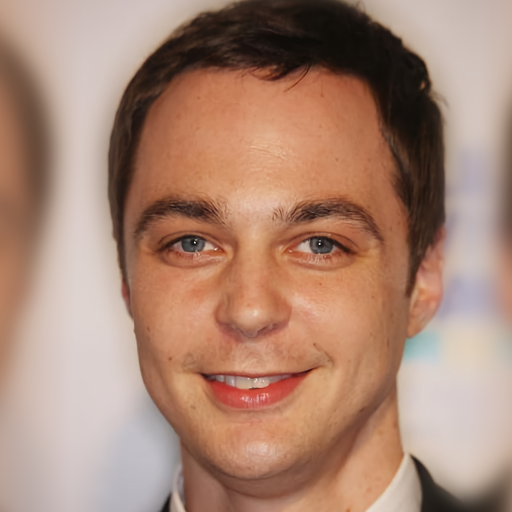} &
\includegraphics[width=\mywidth]{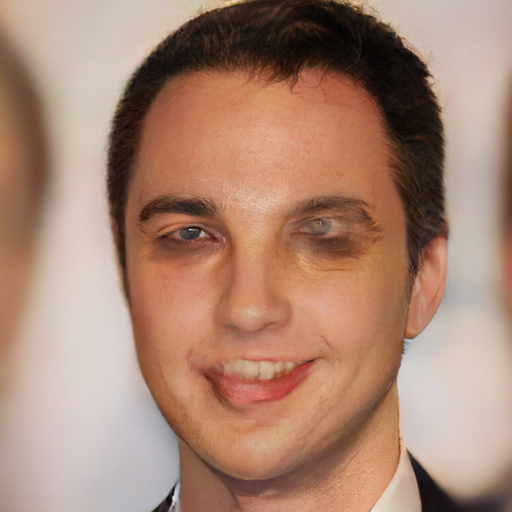} &
\includegraphics[width=\mywidth]{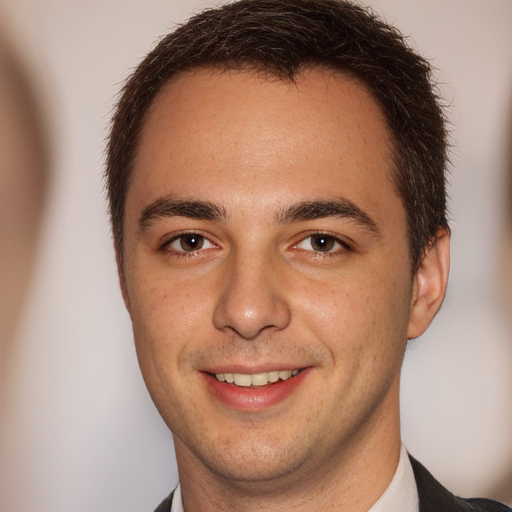} &
\includegraphics[width=\mywidth]{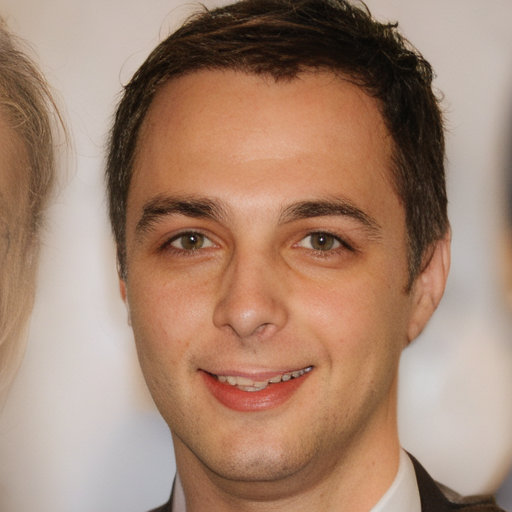} &
\includegraphics[width=\mywidth]{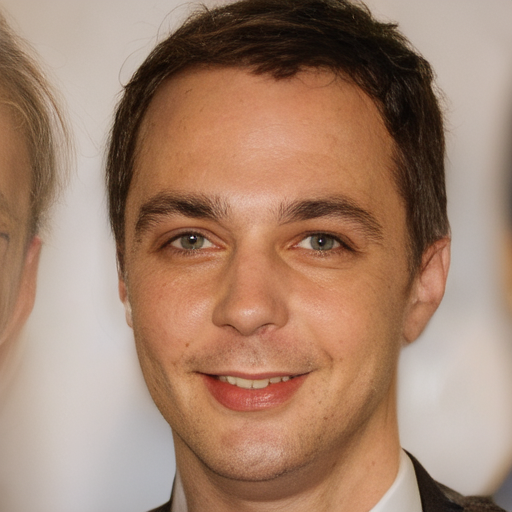} \\
Input LQ & GT & DMDNet & CodeFormer & LDM & \ourmodel \\
\end{tabular}
\caption{
Qualitative comparison. From left to right: input LQ, ground truth, other methods, and our \ourmodel. From top to bottom: \ffhqrefsevere, \ffhqrefmoderate, and CelebA-Test-Ref.
}
\label{fig:qualitative_comparison}
\end{figure}

In Fig.~\ref{fig:qualitative_comparison}, we present a qualitative comparison between our \ourmodel, the pre-trained LDM without reference images, CodeFomer (a SOTA non-reference-based method), and DMDNet (a SOTA reference-based method).
Given the severely degraded image, DMDNet generates distorted face images based on incorrectly detected landmarks.
While CodeFormer yields realistic face images, it does not preserve the facial identity well.
In contrast, our \ourmodel produces results that are both realistic and faithful to the individual's facial identity.

\section{Limitations}
\label{sec:limitation}
When the face region is occluded by other objects, our model may generate artifacts.
For certain face poses (e.g, side face), the reconstructed eyes may appear unnatural.
These problems are also commonly observed in other methods and might be caused due to the lack of such training images.
However, there are some examples showing that these problems can be alleviated if our model is provided reference images with similar face poses to the target image.
Visual examples of these limitations are provided in Appendix~\ref{sec:appendix_vis_limitation}.

\section{Conclusion}
In summary, we propose \ourmodel, which incorporates the \cachekv mechanism and the \timeidloss, to effectively utilize multiple reference images for face restoration.
Additionally, we construct the \ffhqref dataset, which surpasses the existing dataset in both quantity and quality, to facilitate the research in reference-based face restoration.
Evaluation results demonstrate that \ourmodel achieves superior performance in face identity similarity over state-of-the-art methods.

\newpage
\large\noindent\textbf{Acknowledgments}\\
\noindent The authors wish to express their gratitude to Professor Wei-Chen Chiu for his valuable suggestion to exclude the reference images that are too similar to the target images when constructing the proposed \ffhqref dataset.

\bibliography{refldm}

\newpage
\appendix
\section{Broader Impacts}
\label{sec:appendix_broader}
The \ourmodel has the capability to leverage personal appearances from reference images. This introduces a potential risk of misuse, where it could be employed for malicious face editing by using a low-quality image in conjunction with reference images from a different individual.

\begin{figure}[!h]
\centering

\begin{subfigure}{.48\textwidth}
  \centering
  \begin{minipage}{.32\linewidth}
    \includegraphics[width=\linewidth]{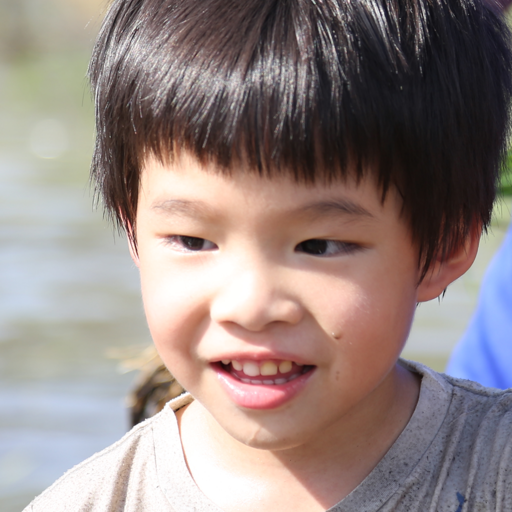}
    \subcaption*{Reference}
  \end{minipage}
  \begin{minipage}{.32\linewidth}
    \includegraphics[width=\linewidth]{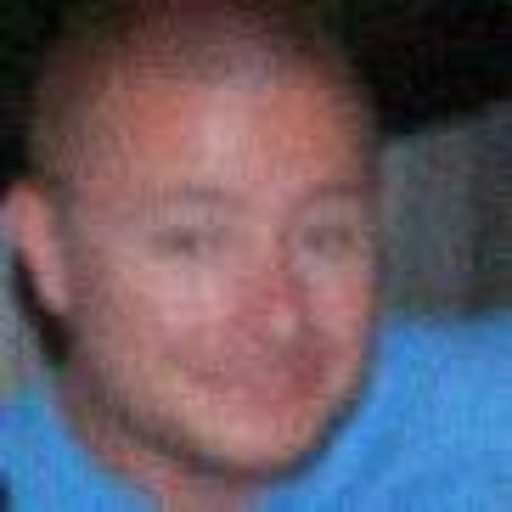}
    \subcaption*{Input LQ}
  \end{minipage}
  \begin{minipage}{.32\linewidth}
    \includegraphics[width=\linewidth]{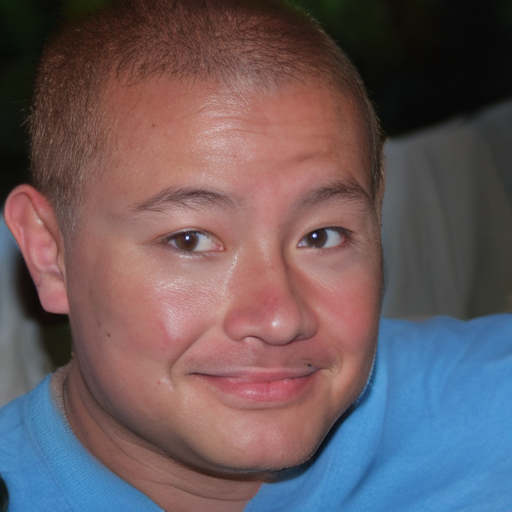}
    \subcaption*{\ourmodel}
  \end{minipage}
  \label{fig:first_set}
\end{subfigure}
\hspace{0.02\textwidth} %
\begin{subfigure}{.48\textwidth}
  \centering
  \begin{minipage}{.32\linewidth}
    \includegraphics[width=\linewidth]{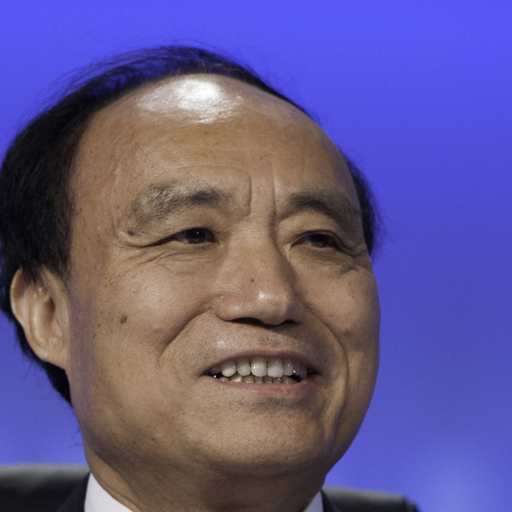}
    \subcaption*{Reference}
  \end{minipage}
  \begin{minipage}{.32\linewidth}
    \includegraphics[width=\linewidth]{fig/diff_person_ref/lq.png}
    \subcaption*{Input LQ}
  \end{minipage}
  \begin{minipage}{.32\linewidth}
    \includegraphics[width=\linewidth]{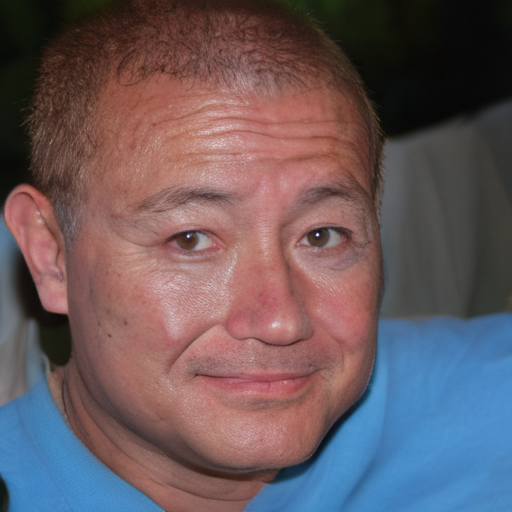}
    \subcaption*{\ourmodel}
  \end{minipage}
  \label{fig:second_set}
\end{subfigure}
\caption{Examples of \ourmodel using reference images from two different individuals.}
\label{fig:comparison_sets}
\end{figure}

\vspace{1em}

\section{Image quality issues in the previous dataset CelebRef-HQ}
\label{sec:appendix_celebrefhq_bad_images}

\begin{figure}[h!]
\centering
\begin{subfigure}{0.19\textwidth}
  \centering
  \includegraphics[width=\linewidth]{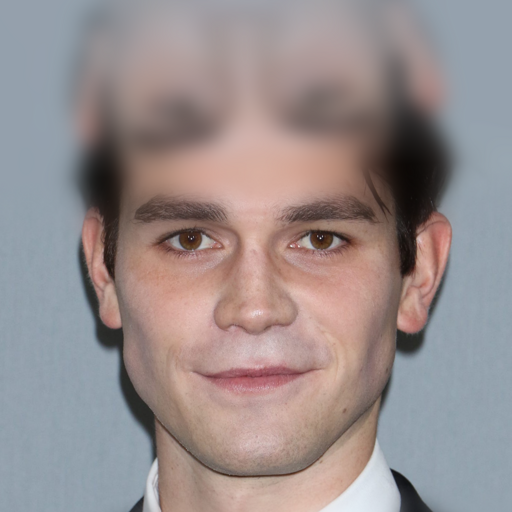}
\end{subfigure}
\hfill
\begin{subfigure}{0.19\textwidth}
  \centering
  \includegraphics[width=\linewidth]{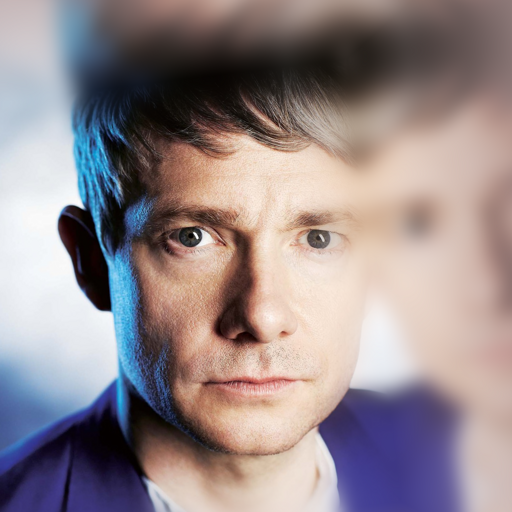}
\end{subfigure}
\hfill
\begin{subfigure}{0.19\textwidth}
  \centering
  \includegraphics[width=\linewidth]{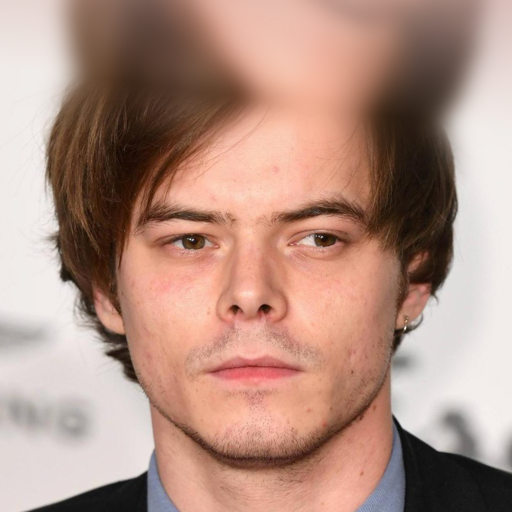}
\end{subfigure}
\hfill
\begin{subfigure}{0.19\textwidth}
  \centering
  \includegraphics[width=\linewidth]{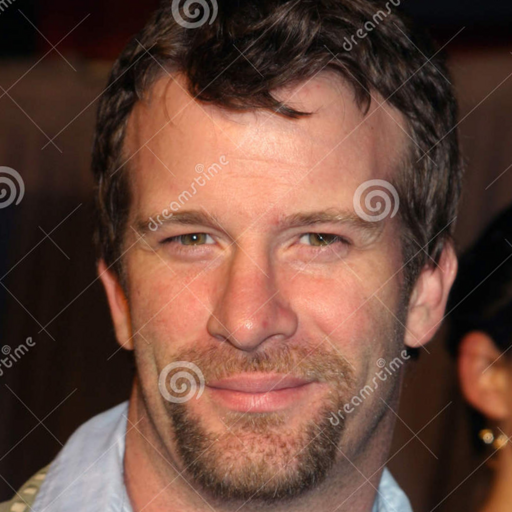}
\end{subfigure}
\hfill
\begin{subfigure}{0.19\textwidth}
  \centering
  \includegraphics[width=\linewidth]{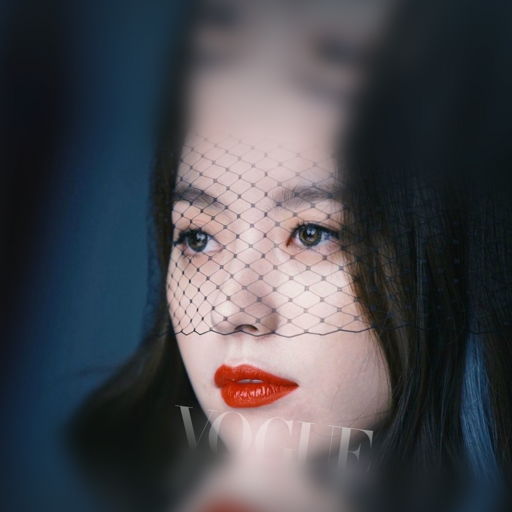}
\end{subfigure}
\caption{Example images with mirror padding and watermark artifacts in the CelebRef-HQ dataset.}
\label{fig:celebrefhq_bad_images}
\end{figure}

As described in Sec.~\ref{subsec:dataset_compare}, the previous dataset for the reference-based face restoration task, CelebRef-HQ~\cite{li2022learning}, exhibits issues with image quality. We provide examples where the ground truth images in this dataset are corrupted by watermarks and mirror padding in Fig.~\ref{fig:celebrefhq_bad_images}.

\vspace{1em}
\section{Statistics of \ffhqref dataset}
\label{sec:appendix_ffhqref_stats}
We analyze the statistics of the proposed \ffhqref dataset, introduced in Sec.~\ref{sec:dataset}.

\begin{figure}[!h]
    \centering
    \begin{subfigure}[b]{0.29\linewidth}
        \includegraphics[width=\linewidth]{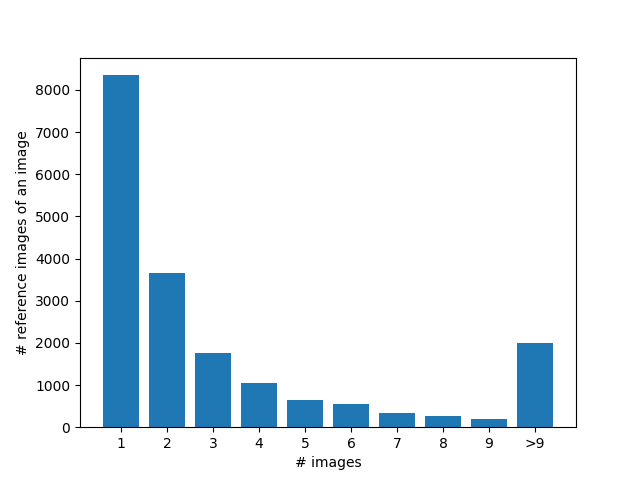}
        \caption{Training set}
        \label{fig:nrefs_per_img_train}
    \end{subfigure}
    \hfill
    \begin{subfigure}[b]{0.29\linewidth}
        \includegraphics[width=\linewidth]{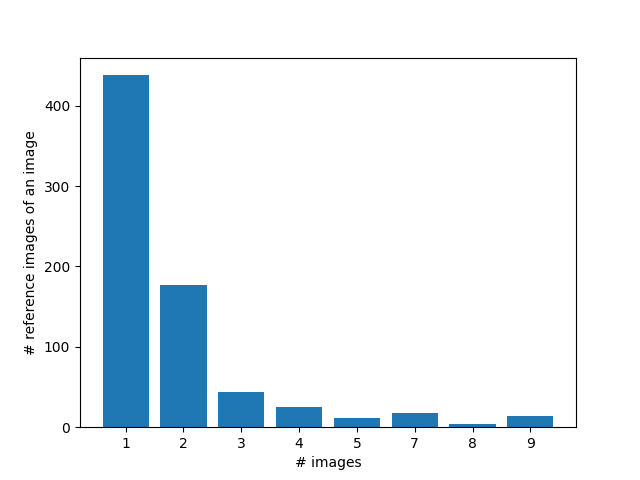}
        \caption{Validation set}
        \label{fig:nrefs_per_img_val}
    \end{subfigure}
    \hfill
    \begin{subfigure}[b]{0.29\linewidth}
        \includegraphics[width=\linewidth]{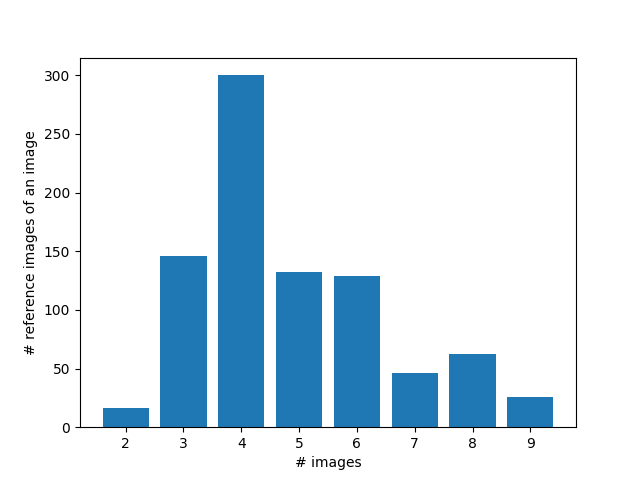}
        \caption{Test set}
        \label{fig:nrefs_per_img_test}
    \end{subfigure}
    \caption{Distribution of the number of available reference images per image in the \ffhqref dataset of train, validation, and test splits.}
    \label{fig:ffhqref_nrefs_per_img}
\end{figure}

In Fig.~\ref{fig:ffhqref_nrefs_per_img}, we plot the distribution of the number of available reference images.

\begin{figure}[!h]
\centering
\begin{subfigure}[b]{0.48\textwidth}
    \centering
    \includegraphics[width=\textwidth]{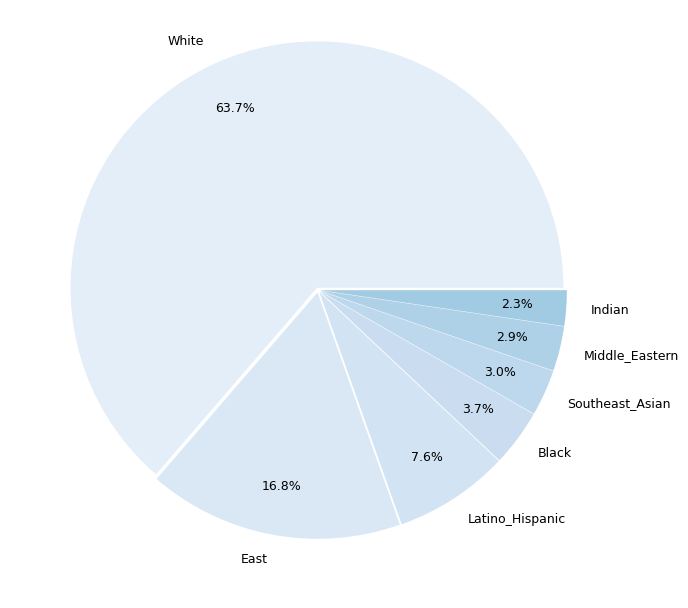}
    \caption{FFHQ-Ref}
    \label{fig:race_ffhqref}
\end{subfigure}
\hfill
\begin{subfigure}[b]{0.48\textwidth}
    \centering
    \includegraphics[width=\textwidth]{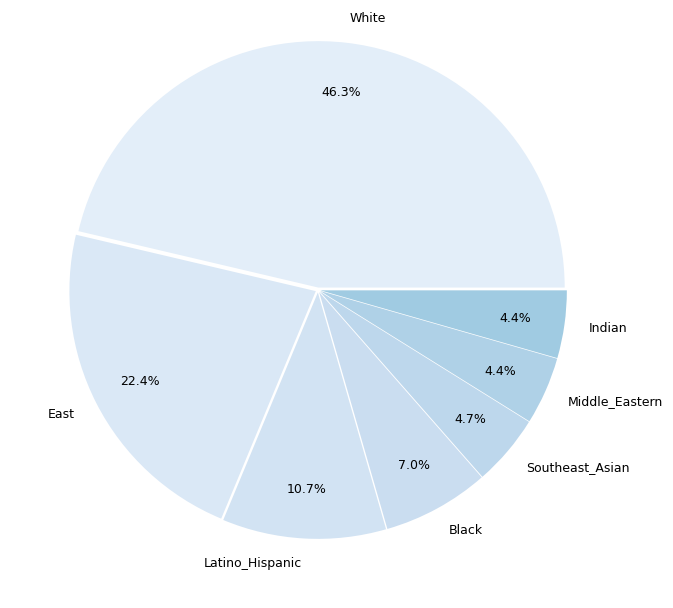}
    \caption{FFHQ-Ref-Test}
    \label{fig:race_ffhqreftest}
\end{subfigure}
\caption{Race distribution within \ffhqref dataset.}
\label{fig:dataset_plot_race}
\end{figure}

\begin{figure}[!h]
\centering
\begin{subfigure}[b]{0.48\textwidth}
    \captionsetup{skip=0pt}
    \centering
    \includegraphics[width=\textwidth]{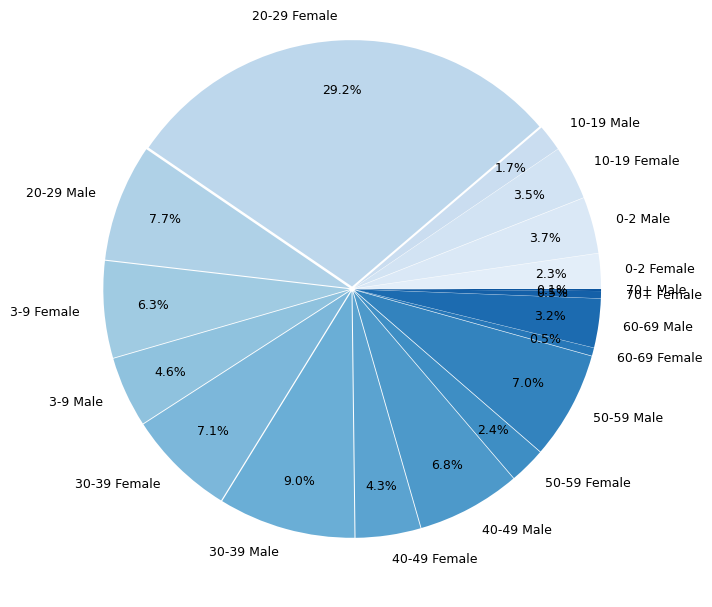}
    \caption{FFHQ-Ref}
    \label{fig:agegender_ffhqref}
\end{subfigure}
\hfill
\begin{subfigure}[b]{0.48\textwidth}
    \captionsetup{skip=0pt}
    \centering
    \includegraphics[width=\textwidth]{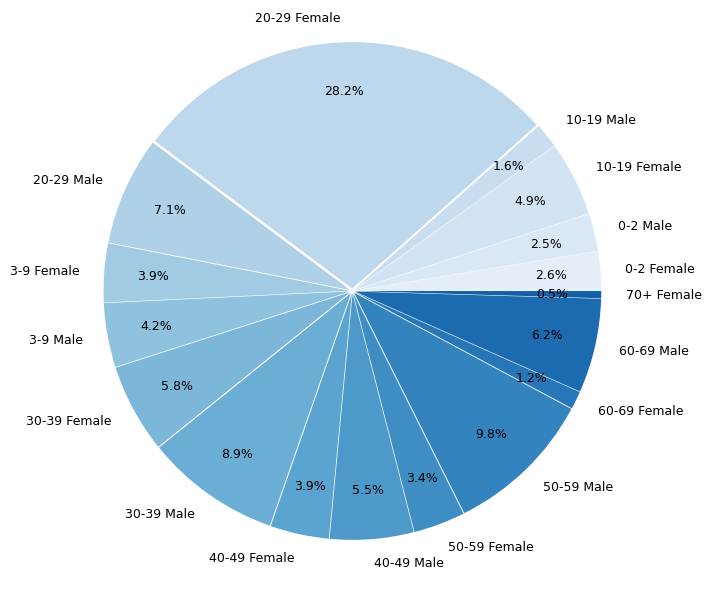}
    \caption{FFHQ-Ref-Test}
    \label{fig:agegender_ffhqreftest}
\end{subfigure}
\caption{Age and gender distribution within \ffhqref dataset.}
\label{fig:dataset_plot_agegender}
\end{figure}

Furthermore, we assess the race, age, and gender distributions of the dataset using labels predicted by FairFace~\cite{karkkainen2021fairface}. As depicted in Fig.~\ref{fig:dataset_plot_race}, the race distribution within \ffhqref is imbalanced, with a predominance of the 'white' category. To mitigate this, we intentionally sampled a greater number of images from other races to construct a more balanced test set. Additionally, as illustrated in Fig.~\ref{fig:dataset_plot_agegender}, \ffhqref encompasses a broad age range, from infants (0-2 years) to the elderly (70+ years). However, the distribution is not uniform across ages and genders. For example, there is a notably higher proportion of young females (29.2\% of '20-29 female').

\section{Examples of differences in background regions}
\label{sec:appendix_background_diff}

\begin{figure}[!h]
\centering
\setlength{\tabcolsep}{1pt} %
\renewcommand{\arraystretch}{0.6} %
\begin{tabular}{cccccc}
\includegraphics[width=0.165\linewidth]{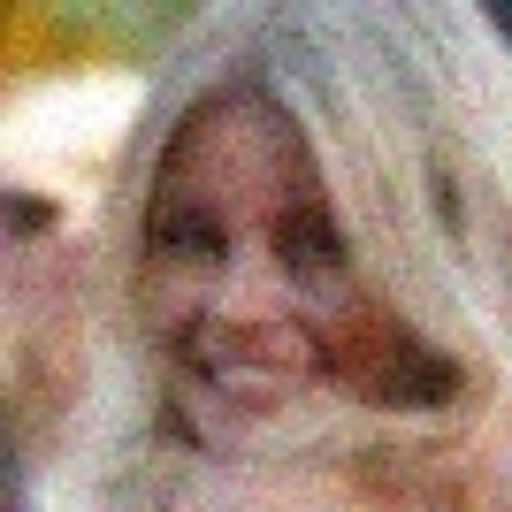} &
\includegraphics[width=0.165\linewidth]{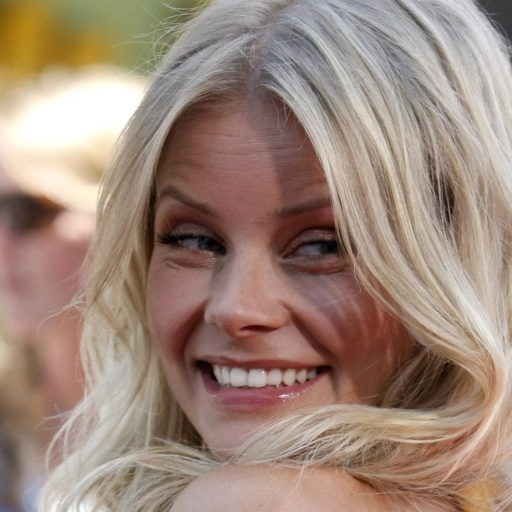} &
\includegraphics[width=0.165\linewidth]{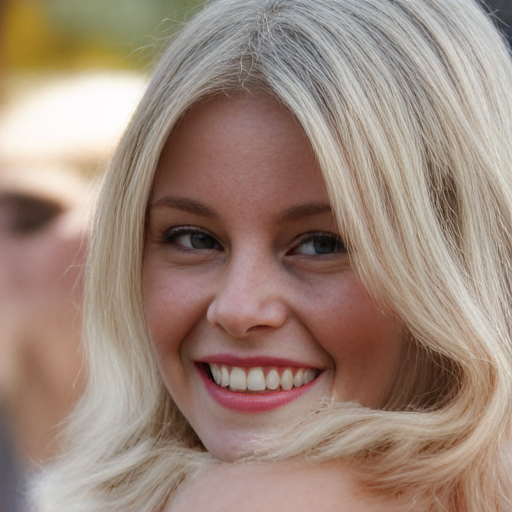} &
\includegraphics[width=0.165\linewidth]{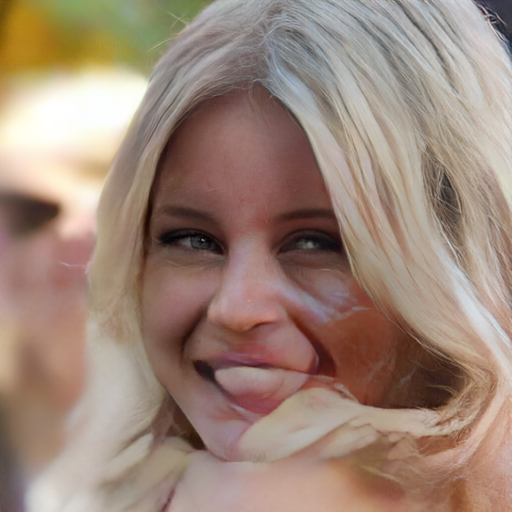} &
\includegraphics[width=0.165\linewidth]{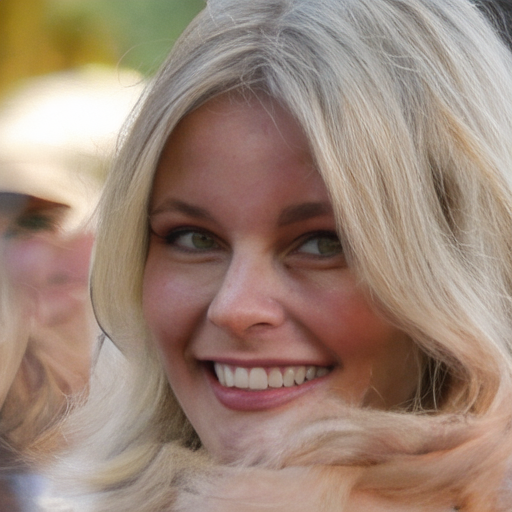} &
\includegraphics[width=0.165\linewidth]{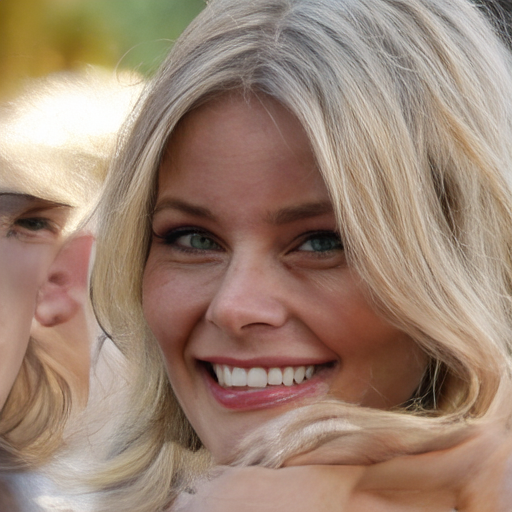} \\
\includegraphics[width=0.165\linewidth]{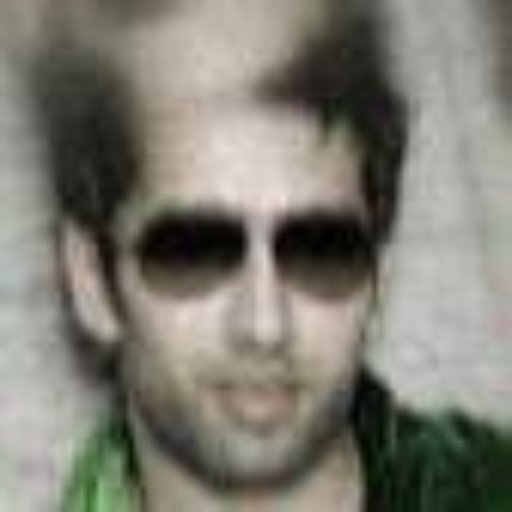} &
\includegraphics[width=0.165\linewidth]{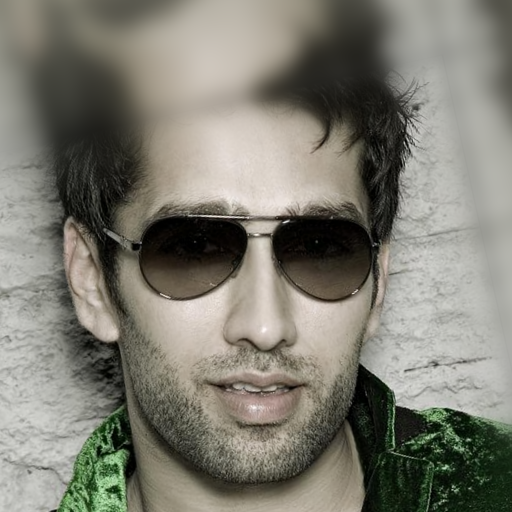} &
\includegraphics[width=0.165\linewidth]{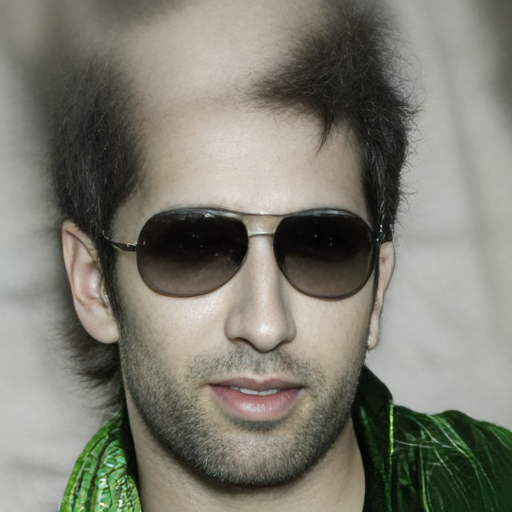} &
\includegraphics[width=0.165\linewidth]{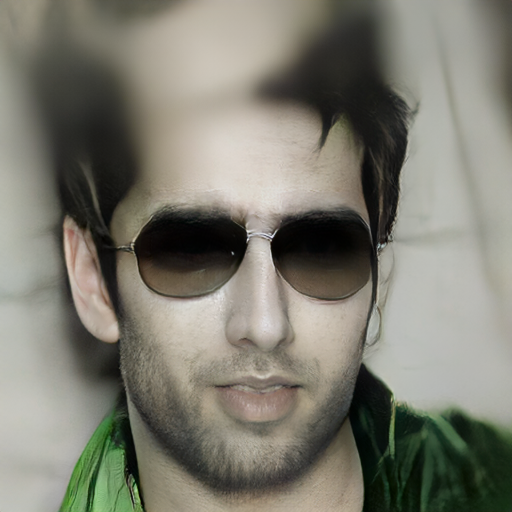} &
\includegraphics[width=0.165\linewidth]{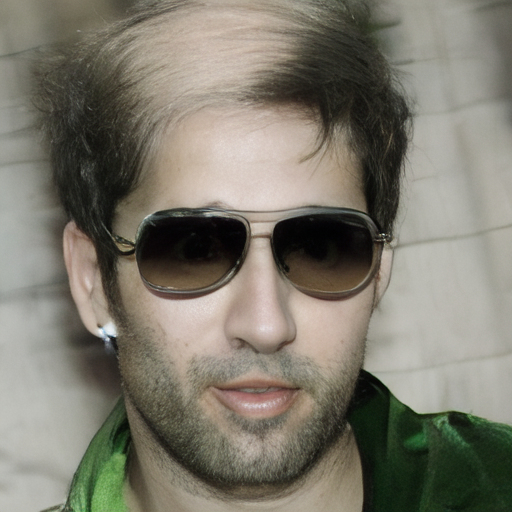} &
\includegraphics[width=0.165\linewidth]{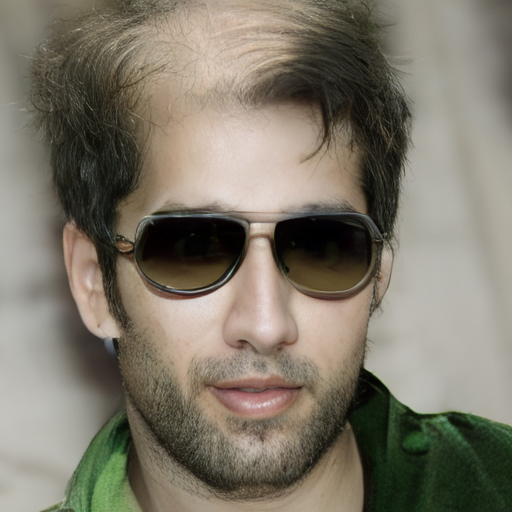} \\

Input LQ & GT & CodeFormer & DMDNet & LDM  & \ourmodel \\
\end{tabular}
\caption{Examples where \ourmodel generates background-region details that differ more from the ground truth.}
\label{fig:background_diff}
\end{figure}

In Fig.~\ref{fig:background_diff}, we provide some examples where our \ourmodel are more different to the ground truth in background pixels compared to prior methods. In the first example, the \ourmodel attempts to restore another face in the background. In the second example, our \ourmodel restored the mirror padding in the CelebA-Test dataset as hairs.

\section{Examples of illumination change}
\label{sec:appendix_illumination_reference}

\begin{figure}[!h]
\centering
\begin{subfigure}[b]{0.24\textwidth}
    \centering
    \includegraphics[width=\textwidth]{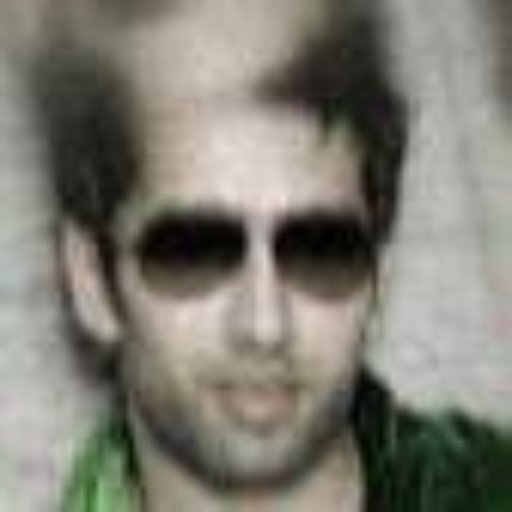}
    \caption{Input LQ}
\end{subfigure}
\hfill
\begin{subfigure}[b]{0.24\textwidth}
    \centering
    \includegraphics[width=\textwidth]{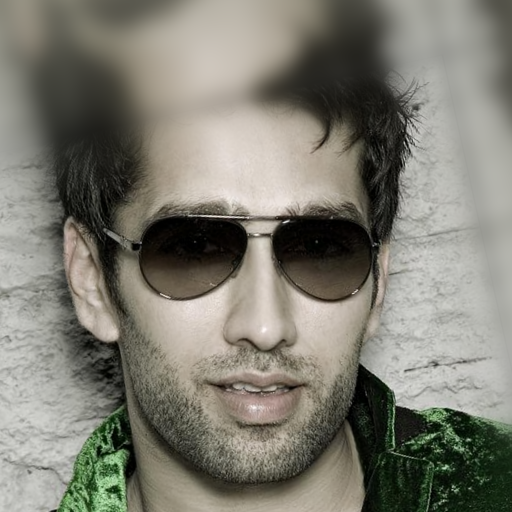}
    \caption{GT}
\end{subfigure}
\hfill
\begin{subfigure}[b]{0.24\textwidth}
    \centering
    \includegraphics[width=\textwidth]{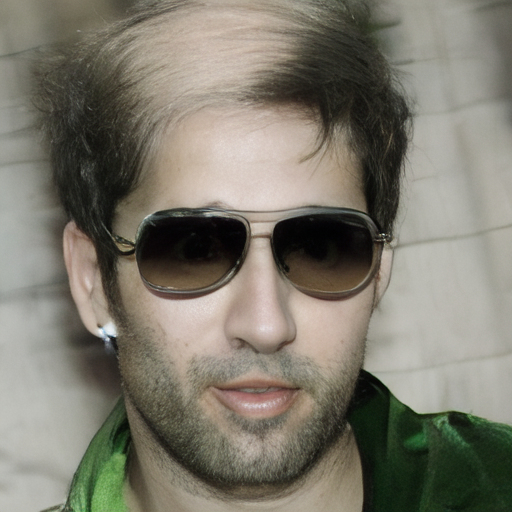}
    \caption{LDM}
\end{subfigure}
\hfill
\begin{subfigure}[b]{0.24\textwidth}
    \centering
    \includegraphics[width=\textwidth]{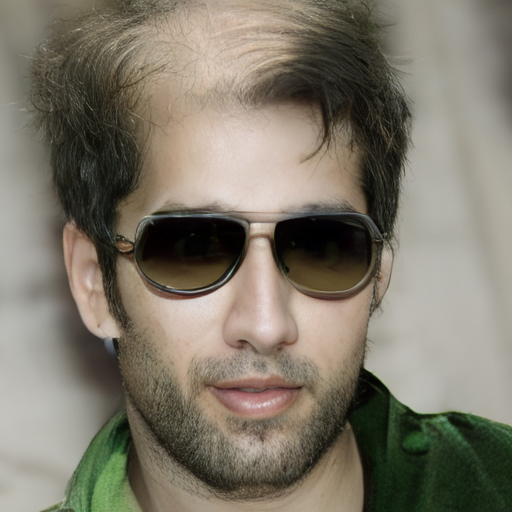}
    \caption{RefLDM}
\end{subfigure}
\begin{subfigure}[b]{1.0\textwidth}
    \centering
    \includegraphics[width=\textwidth]{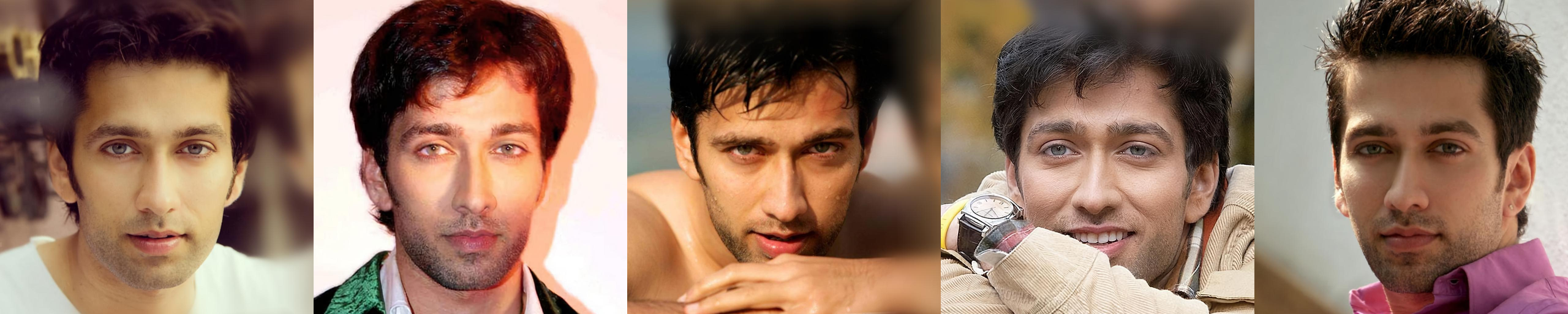}
    \caption{Input references}
\end{subfigure}
\caption{An example of (d) ReF-LDM demonstrating an illumination change, likely influenced by the strong warm lighting of the (e) input reference images.}
\label{fig:illumination_reference}
\end{figure}

Fig.~\ref{fig:illumination_reference} shows an example where \ourmodel exhibits warmer illumination compared to LDM. We conjecture that this may be due to the impact of the strong warm lighting in the input reference images. To address this issue, one could employ post-processing tricks, such as adjusting the means of the R, G, B channels to match those of the input LQ image. Another potential solution might be training \ourmodel with data augmentation on the illuminations of input reference images, to encourage the model to disregard the illuminations of input references and maintain consistency with that of the input LQ image.

\section{Examples of failure cases}
\label{sec:appendix_vis_limitation}

\begin{figure}[!ht]
\centering
\setlength{\tabcolsep}{1pt} %
\renewcommand{\arraystretch}{0.6} %
\begin{tabular}{cccccc}
\includegraphics[width=0.165\linewidth]{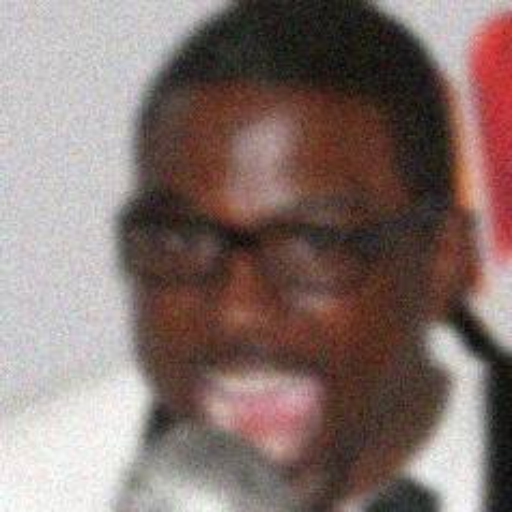} &
\includegraphics[width=0.165\linewidth]{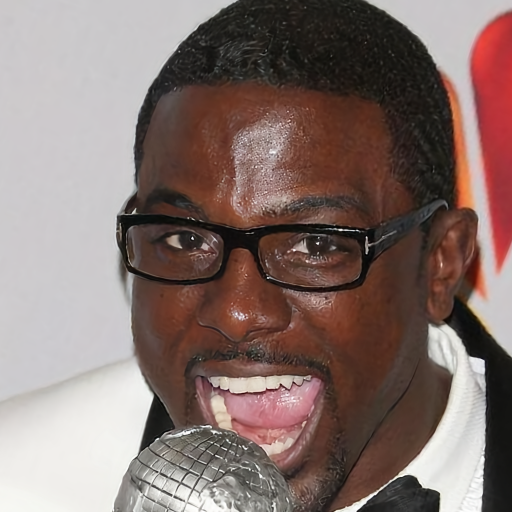} &
\includegraphics[width=0.165\linewidth]{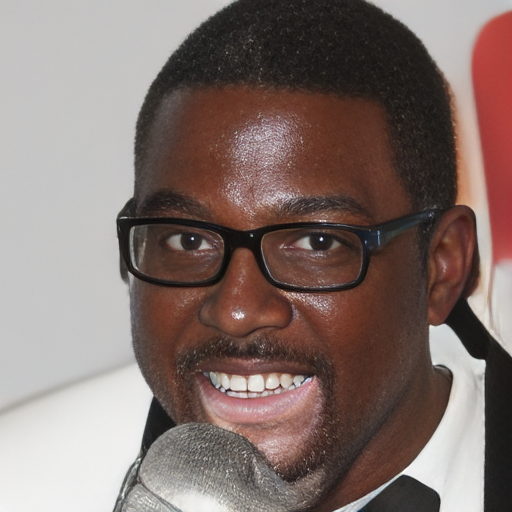} &
\includegraphics[width=0.165\linewidth]{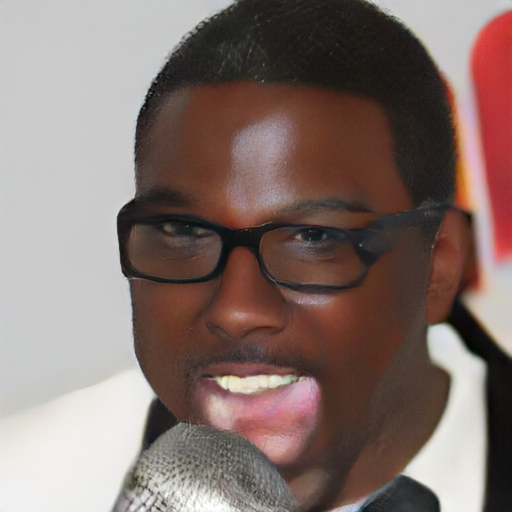} &
\includegraphics[width=0.165\linewidth]{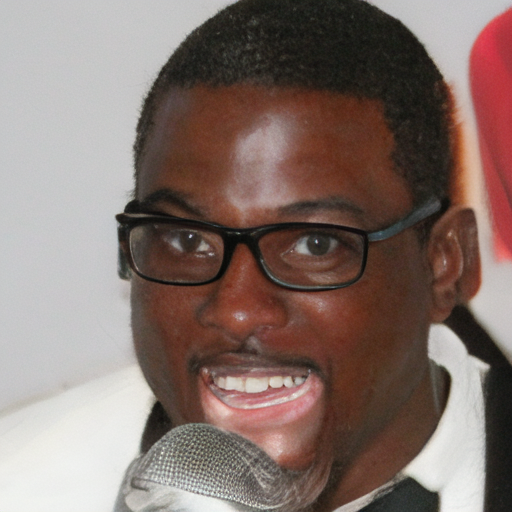} &
\includegraphics[width=0.165\linewidth]{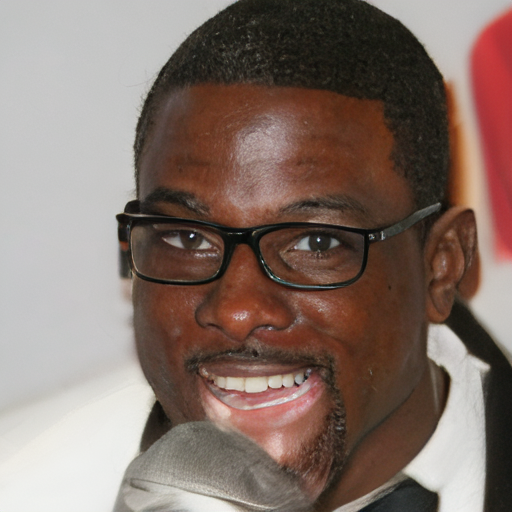} \\
\includegraphics[width=0.165\linewidth]{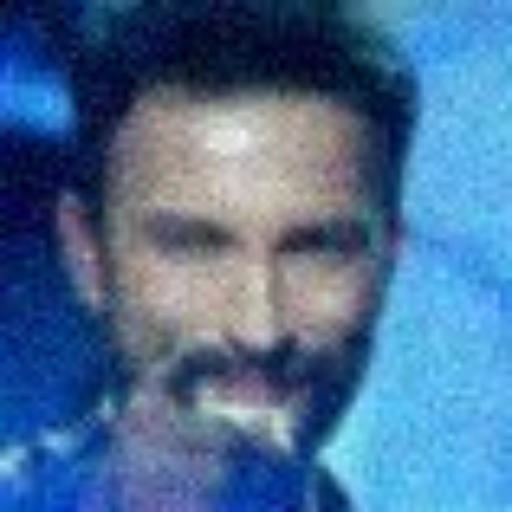} &
\includegraphics[width=0.165\linewidth]{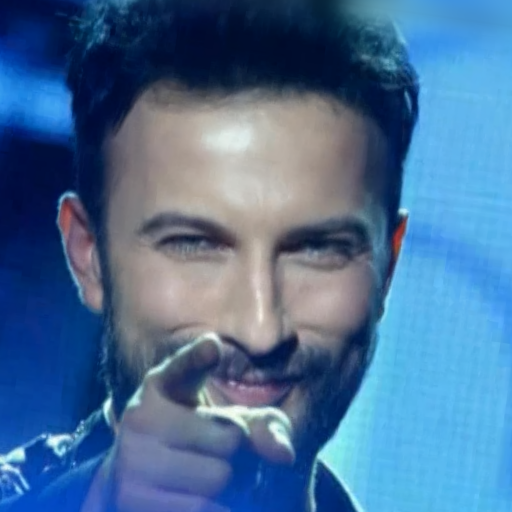} &
\includegraphics[width=0.165\linewidth]{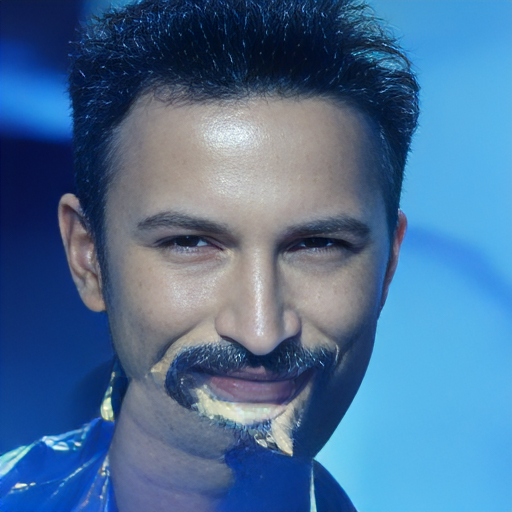} &
\includegraphics[width=0.165\linewidth]{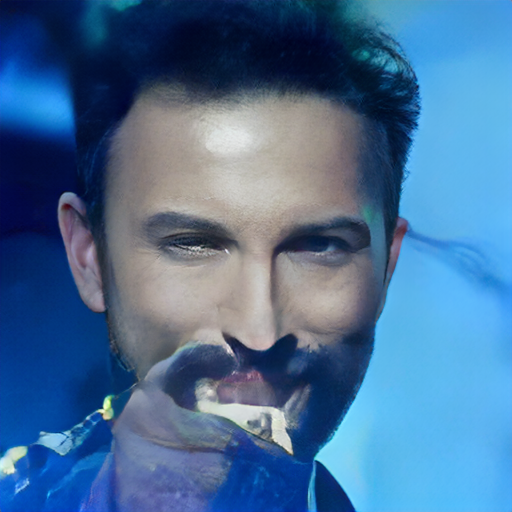} &
\includegraphics[width=0.165\linewidth]{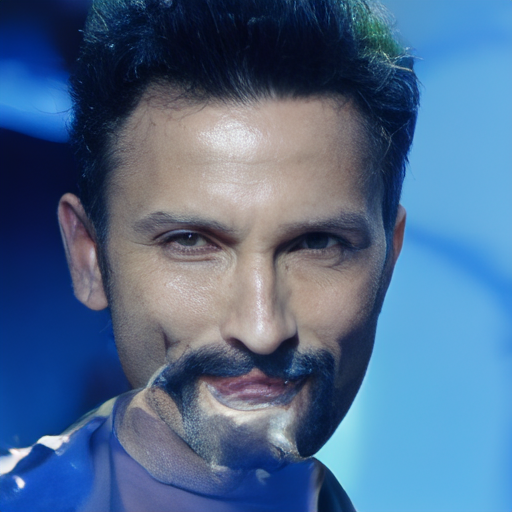} &
\includegraphics[width=0.165\linewidth]{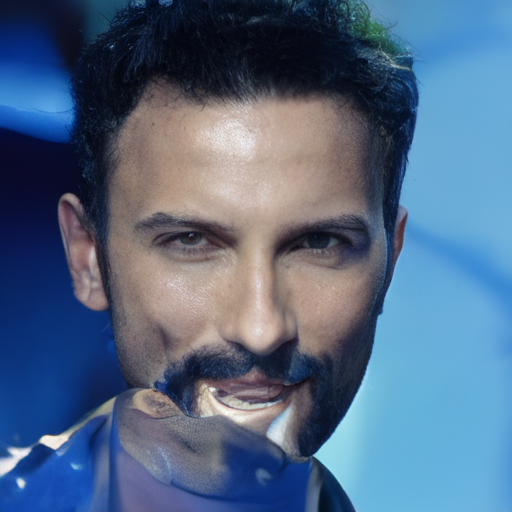} \\
Input LQ & GT & CodeFormer & DMDNet & LDM  & \ourmodel \\
\end{tabular}
\caption{
Some failure cases when the input LQ images are occluded.
}
\label{fig:limitation_occlusion}
\end{figure}

\begin{figure}[!h]
\centering
\setlength{\tabcolsep}{1pt} %
\renewcommand{\arraystretch}{0.6} %
\begin{tabular}{cccccc}
\includegraphics[width=0.165\linewidth]{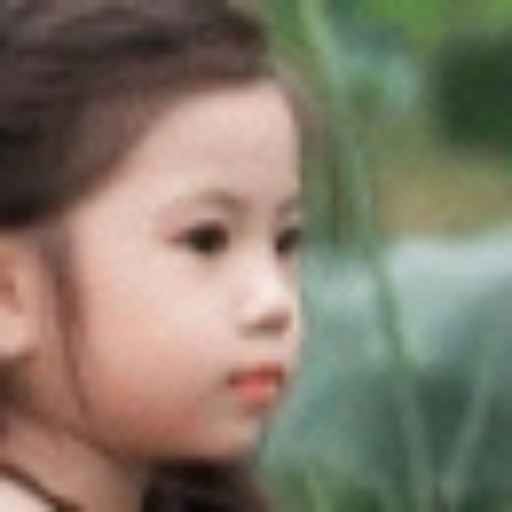} &
\includegraphics[width=0.165\linewidth]{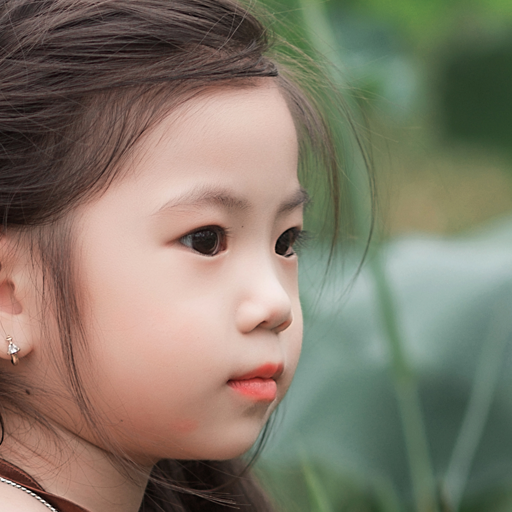} &
\includegraphics[width=0.165\linewidth]{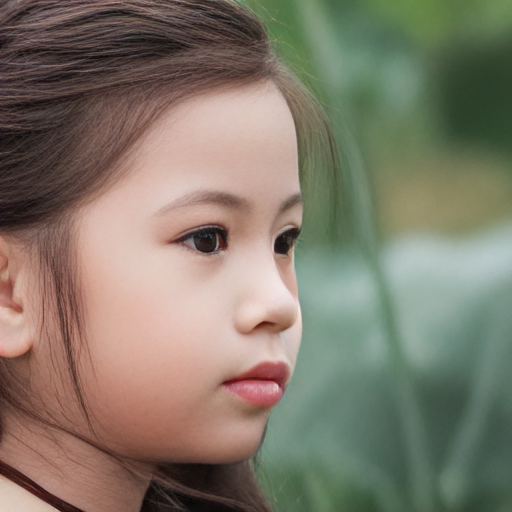} &
\includegraphics[width=0.165\linewidth]{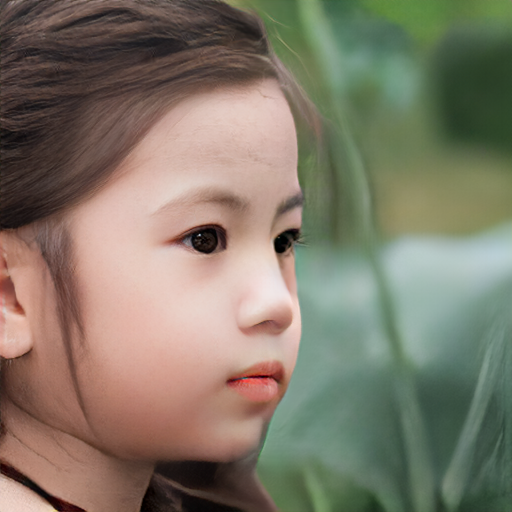} &
\includegraphics[width=0.165\linewidth]{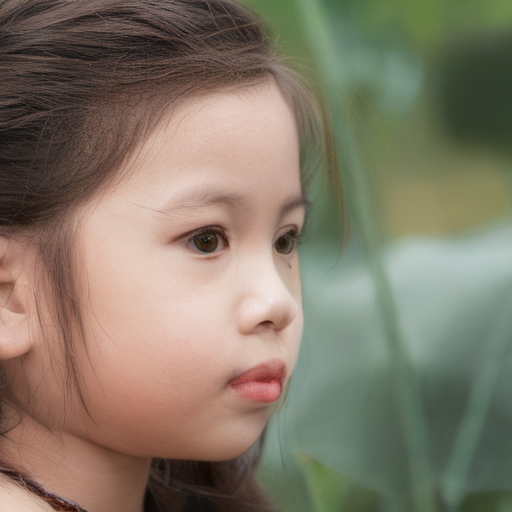} &
\includegraphics[width=0.165\linewidth]{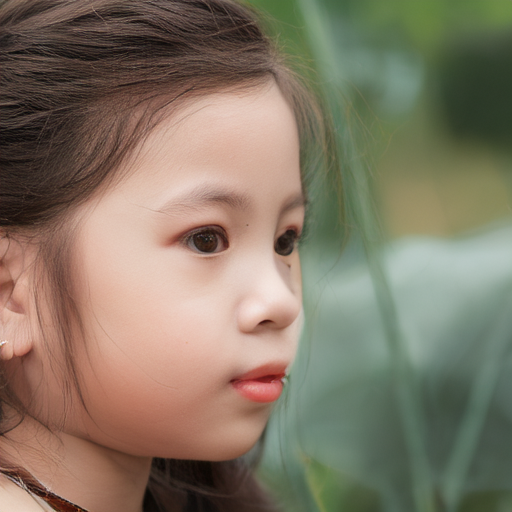} \\
\includegraphics[width=0.165\linewidth]{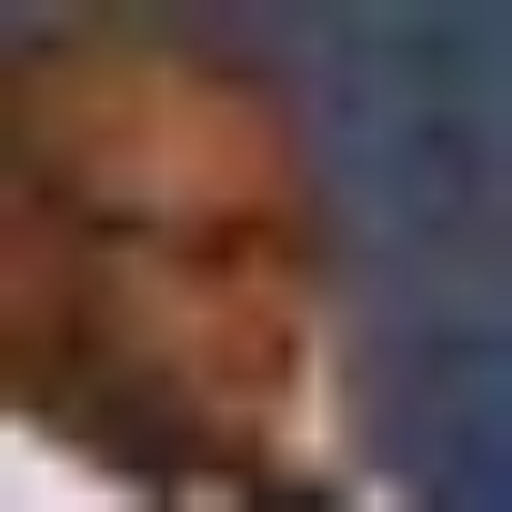} &
\includegraphics[width=0.165\linewidth]{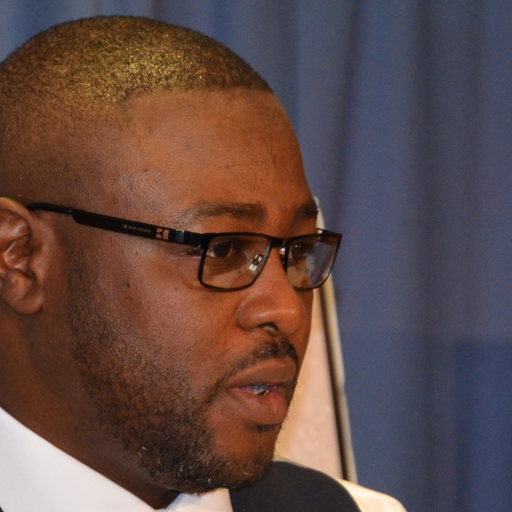} &
\includegraphics[width=0.165\linewidth]{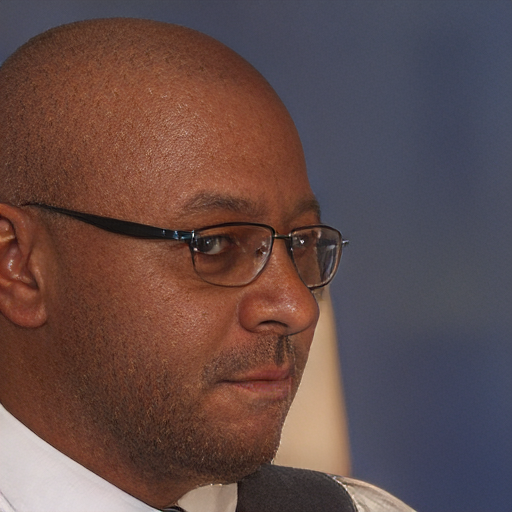} &
\includegraphics[width=0.165\linewidth]{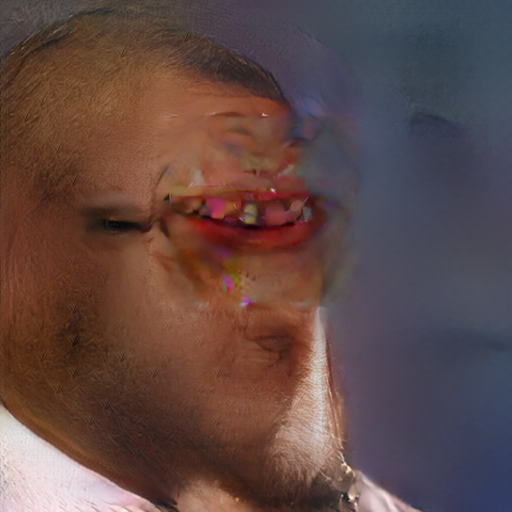} &
\includegraphics[width=0.165\linewidth]{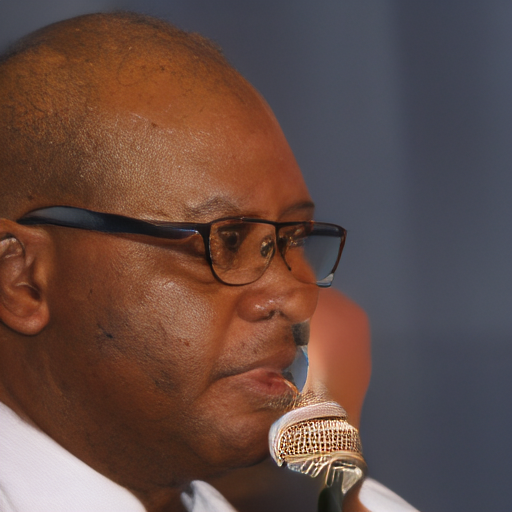} &
\includegraphics[width=0.165\linewidth]{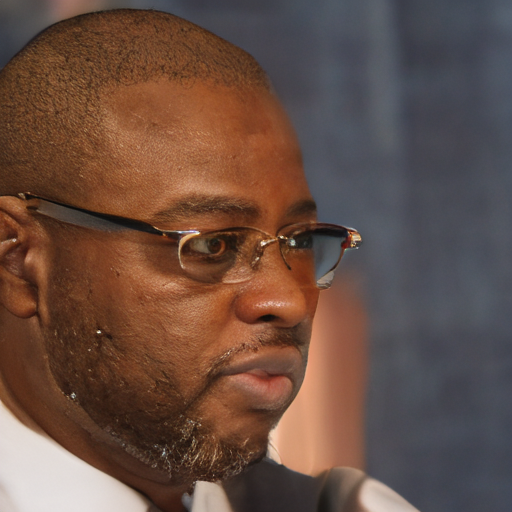} \\
Input LQ & GT & CodeFormer & DMDNet & LDM  & \ourmodel \\
\end{tabular}
\caption{Some failure cases of side faces when side-face images are absent in the references.}
\label{fig:limitation_sideface_wo_sideref}
\end{figure}

\begin{figure}[!h]
\centering
\setlength{\tabcolsep}{1pt} %
\renewcommand{\arraystretch}{0.6} %
\begin{tabular}{cccccc}
\includegraphics[width=0.165\linewidth]{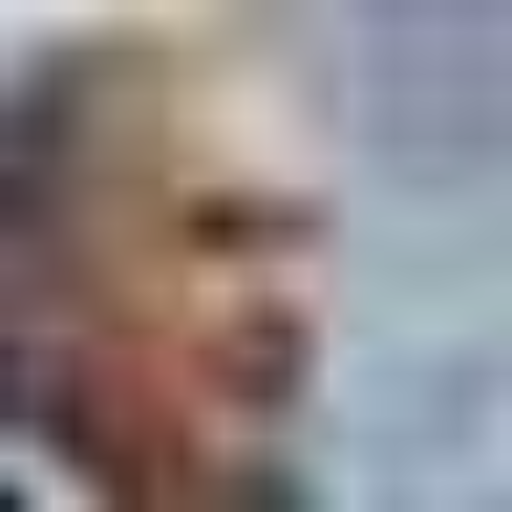} &
\includegraphics[width=0.165\linewidth]{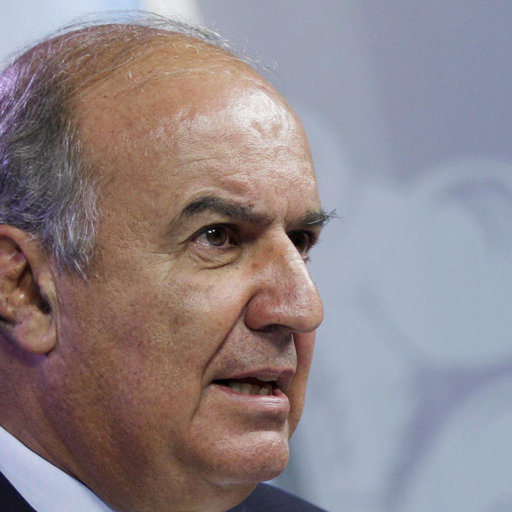} &
\includegraphics[width=0.165\linewidth]{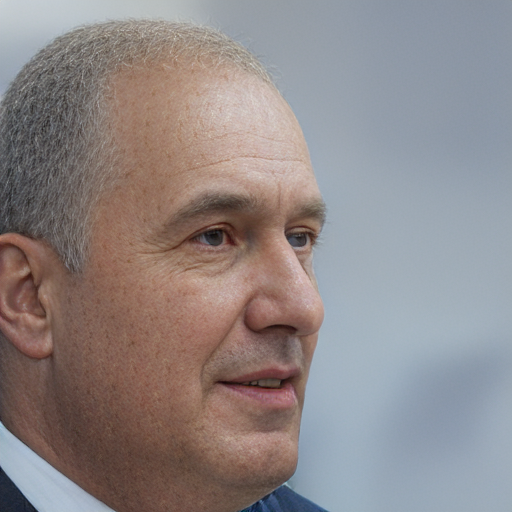} &
\includegraphics[width=0.165\linewidth]{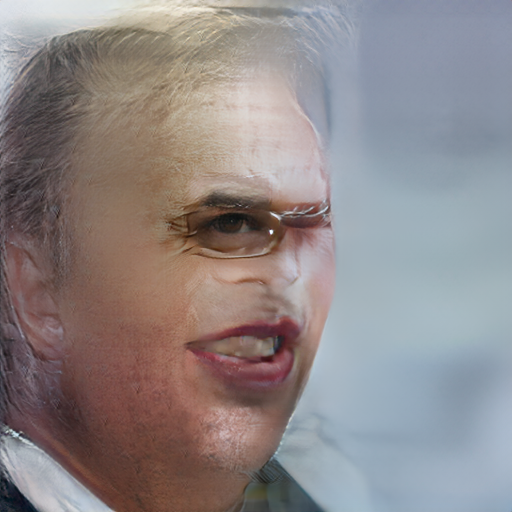} &
\includegraphics[width=0.165\linewidth]{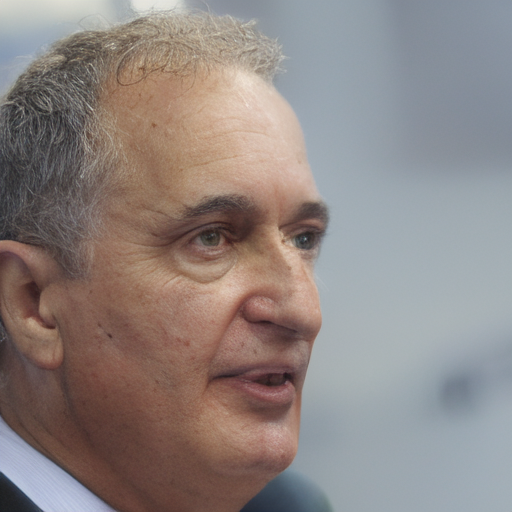} &
\includegraphics[width=0.165\linewidth]{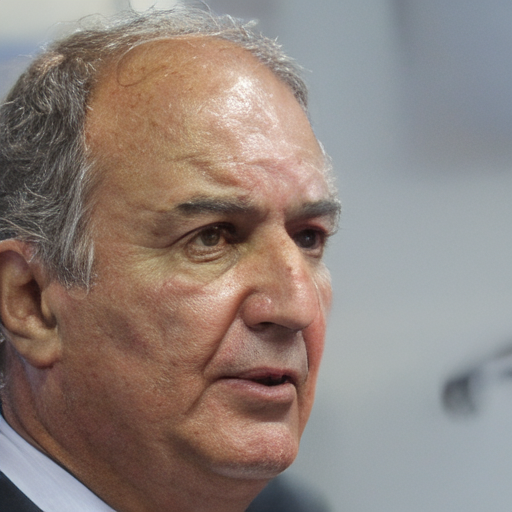} \\
\includegraphics[width=0.165\linewidth]{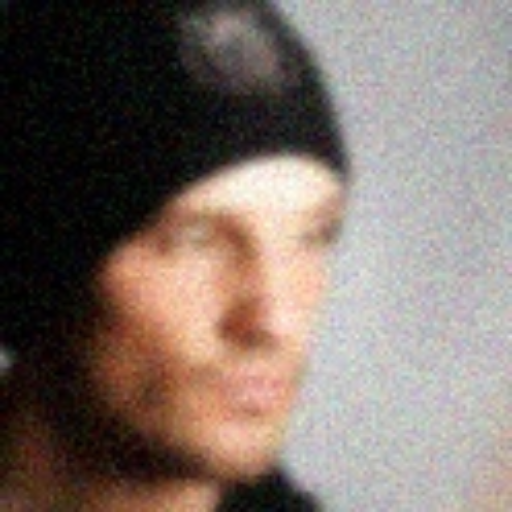} &
\includegraphics[width=0.165\linewidth]{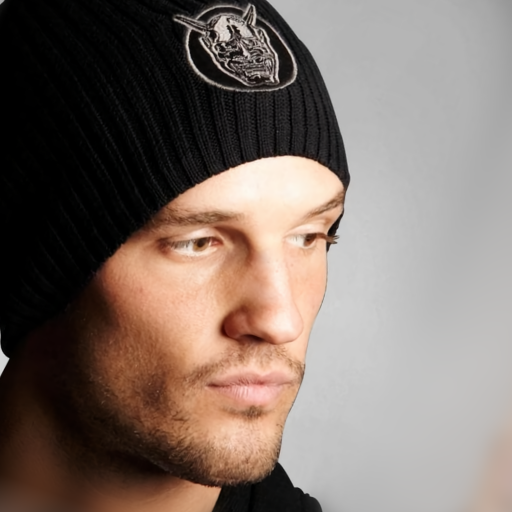} &
\includegraphics[width=0.165\linewidth]{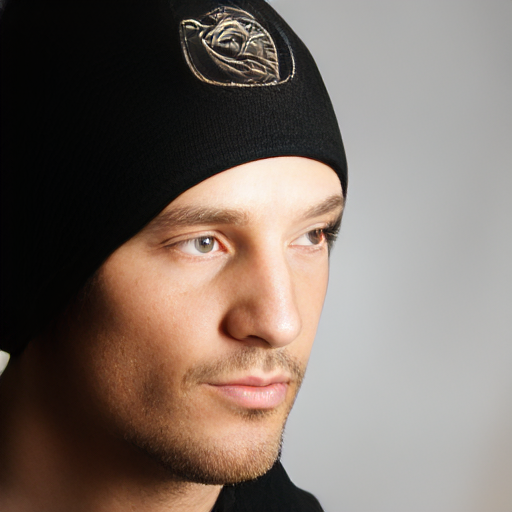} &
\includegraphics[width=0.165\linewidth]{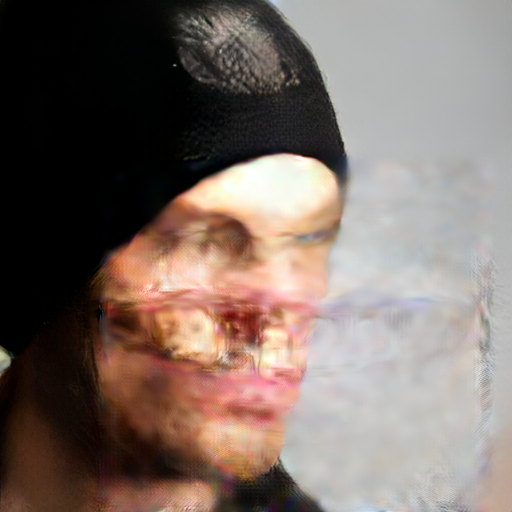} &
\includegraphics[width=0.165\linewidth]{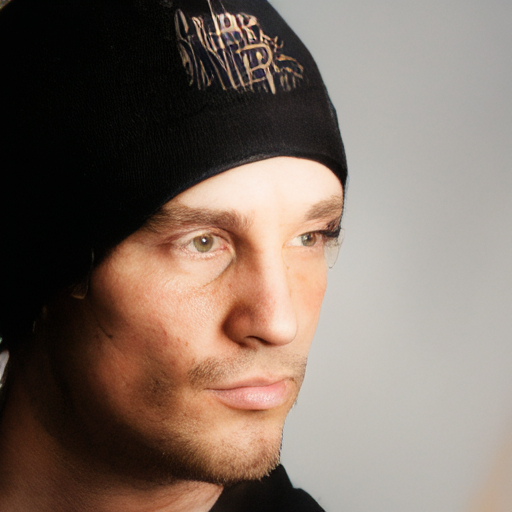} &
\includegraphics[width=0.165\linewidth]{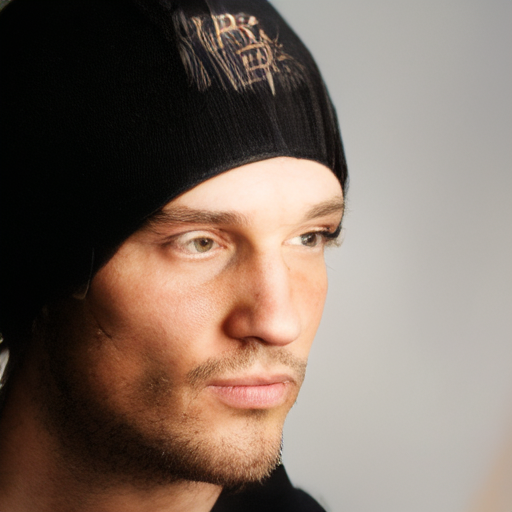} \\
Input LQ & GT & CodeFormer & DMDNet & LDM  & \ourmodel \\
\end{tabular}
\caption{Some successful cases of side faces when side-face images are included in the references.}
\label{fig:limitation_sideface_w_sideref}
\end{figure}

Here we provide visual examples for the limitation described in Sec.~\ref{sec:limitation}.
As shown Fig.~\ref{fig:limitation_occlusion}, when the face region is occluded, our \ourmodel and the prior models tend to generate from unnatural artifacts.
For side face images, the \ourmodel may not work well when the input reference images do not contain faces of similar pose, as shown in Fig.~\ref{fig:limitation_sideface_wo_sideref}.
However, Fig.~\ref{fig:limitation_sideface_w_sideref} suggests that our \ourmodel can effectively exploit the reference images of similar face poses to improve the results.

\section{More implementation details}
\label{sec:appendix_more_impl}

\subsection{Classifier-free guidance towards reference images}
\label{subsec:appendix_cfg_ref}
Classifier-free guidance~\cite{ho2022classifier} is a technique widely used in diffusion models for guiding the generated results towards a condition $c$ with a controllable scale factor $s$ at inference time:
\begin{equation}
\tilde{\epsilon}_\theta(\zt, c) = \epsilon_\theta(\zt, \varnothing) + s \cdot \left( \epsilon_\theta(\zt, c) - \epsilon_\theta(\zt, \varnothing) \right)
\end{equation}

In our experiments, we use classifier-free guidance towards reference images with $s=1.5$.
\begin{equation}
\tilde{\epsilon}_\theta(\zt, \zlq, \zrefs) = \epsilon_\theta(\zt, \zlq, \varnothing) + s \cdot \left( \epsilon_\theta(\zt, \zlq, \zrefs) - \epsilon_\theta(\zt, \zlq, \varnothing) \right)
\end{equation}
During the training phase, we randomly drop the conditions by setting them to zero tensors with a probability of 0.1.

\subsection{Data augmentation for input reference images}
\label{subsec:appendix_data_aug_ref}
\begin{figure}[!ht]
    \centering
    \includegraphics[width=\linewidth]{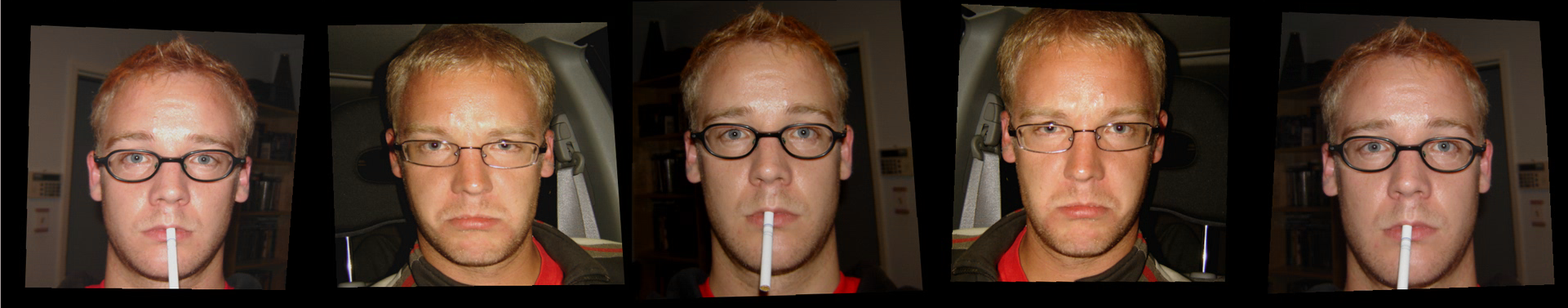}
    \caption{Data augmentation for reference images.}
    \label{fig:data_aug_ref}
\end{figure}

During the training phase, we use a fixed number of five input reference images. When a target images with less than five reference images are sampled, we repeat the reference images to obtain five reference images. In addition, we apply image augmentation to the input reference images with the following operations: color jitter (brightness ± 0.2, contrast ±  0.2, saturation ±  0.2, hue ±  0.02), affine transform (rotation ± 2, translation ± 0.05, scale ± 0.05), perspective transform (scale ± 0.2, probability 0.5), and horizontal flip (probability 0.5). Lastly, we randomly shuffle the order of available reference images for a target image, so that a different combination of reference images can be sampled at each training iteration. In Fig.~\ref{fig:data_aug_ref}, we provide an example where a set of two reference images is augmented to a set of five reference images.

\subsection{Training details}
\label{sec:appendix_train_resources}
We trained the VQGAN for 200,000 iterations with batch size 32 on four A6000 GPUs for 7 days.
We trained the LDM with only LQ condition for 500,000 iterations with batch size 40 on four A6000 GPUs for 7 days.
We finetuned the \ourmodel for 150,000 iterations with batch size 8 on four 3090 GPUs for 6 days.
For training losses, the LDM is trained using only the typical LDM loss $\lldm$, while the \ourmodel is trained with both $\lldm$ and the proposed $\ltimeid$.

\subsection{Hyperparameters of networks}
\label{subsec:appendix_ldm_arch}
For the frozen autoencoder, we use a VQGAN as in the LDM~\cite{rombach2022high} with the following settings: 
\begin{itemize} 
    \item input image: 512x512x3
    \item latent representation: 64x64x8
    \item code booksize: 8192
    \item network hyperparameters: base channel as 128, multiplier for each scale as [1, 1, 2, 4] with 2 residual blocks.
\end{itemize}

For the denoising U-net, we use the following settings:
\begin{itemize} 
    \item input latent: 64x64x16
    \item output latent: 64x64x8
    \item attention layer at resolutions: 32x32, 16x16, and 8x8
    \item network hyperparameters: base channel as 160, multiplier for each scale as [1, 2, 2, 4] with 2 residual blocks.
\end{itemize}

\end{document}